\documentclass[times, review, 10pt]{elsarticle}
\usepackage{float}
\usepackage{graphicx}
\usepackage{longtable}
\usepackage{tabularx}
\usepackage{xcolor}
\newif\ifclean
\cleantrue          
\ifclean
    \colorlet{red}{black}
\fi
\usepackage{algorithm}
\usepackage{algorithm}
\usepackage{algorithmic}
\usepackage{subcaption}  
\usepackage{booktabs} 
\usepackage{caption} 
\usepackage{setspace} 
\usepackage[mathlines,switch]{lineno}
\usepackage{hyperref}
\usepackage{pifont}
\usepackage{amsmath}
\usepackage{amssymb}

\floatname{algorithm}{Algorithm.}
\usepackage{makecell}
\usepackage{colortbl} 
\usepackage{multirow}
\definecolor{bestcell}{HTML}{FBC8CE}  
\definecolor{secondcell}{HTML}{DADAFE} 
\newcommand{\best}[1]{\cellcolor{bestcell} #1}      
\newcommand{\second}[1]{\cellcolor{secondcell} #1}  
\definecolor{skyblue}{RGB}{16, 127, 172}
\hypersetup{
    colorlinks=true,
    linkcolor=skyblue,
    citecolor=skyblue,
    urlcolor=skyblue
}
\def\tsc#1{\csdef{#1}{\textsc{\lowercase{#1}}\xspace}}
\tsc{WGM}
\tsc{QE}
\tsc{EP}
\tsc{PMS}
\tsc{BEC}
\tsc{DE}
\begin{document} 
\let\WriteBookmarks\relax
\def\floatpagepagefraction{1}
\def\textpagefraction{.001}
\renewcommand{\arraystretch}{1.5} 
\title{Exposure-aware Progressive Optimization Method for Infrared and Visible Image Fusion}
\author[1,2]{Zhiwei Wang}

\author[1]{Defeng He\corref{cor1}}
\author[3]{Li Zhao\corref{cor1}}

\author[1]{Xiaoqin Zhang}

\author[2]{Yuxing Li}

\author[2]{Edmund Y. Lam}
\cortext[cor1]{Corresponding author.}
\affiliation[1]{organization={College of Information Engineering, Zhejiang University of Technology},
    city={Hangzhou},
    postcode={310023}, 
    country={China}}
\affiliation[2]{organization={Department of Electrical and Computer Engineering,  The University of Hong Kong},
    city={Hong Kong},
    country={China}}
\affiliation[3]{organization={College of Information Science and Technology, Zhejiang Shuren University},
    city={Hangzhou},
    postcode={310015}, 
    country={China}}

\begin{abstract}
Overexposure caused by strong daylight and oncoming headlights frequently overwhelms visible sensors, resulting in critical information loss in visual perception. Infrared and visible image fusion can compensate for such degradation via multimodal complementarity. However, most fusion methods lack region-aware optimization for overexposed areas and cannot effectively exploit infrared cues in saturated regions, resulting in insufficient infrared detail preservation or redundant information in the fused results. To address this, we propose EPOFusion, an exposure-aware fusion framework. It employs a spatial guidance module to identify regions requiring infrared compensation, together with a region-aware fusion loss to strengthen informative infrared structures. In addition, an iterative \textcolor{red}{feature} refinement head equipped with a multiscale context fusion module progressively refines fused representations, enabling effective integration of complementary infrared information while maintaining visual consistency in normally exposed regions. The infrared and visible overexposure (IVOE) dataset consists of a synthetic training subset providing \textcolor{red}{infrared-compensation supervision and a real-world subset for fusion and downstream perception evaluation under authentic overexposure.} 
\textcolor{red}{EPOFusion demonstrates superior VIF and $Q^{AB/F}$ performance with favorable visual quality, improving $Q^{AB/F}$ by 10.7\% over the existing overexposure-oriented fusion baseline, while further improving downstream mIoU and mAP50 by 5.6\% and 6.5\%, respectively.}
Code, results, and the IVOE dataset will be made available at \url{https://warren-wzw.github.io/EPOFusion/}.

\end{abstract}

\makeatletter
\def\ps@pprintTitle{%
  \let\@oddhead\@empty
  \let\@evenhead\@empty
  \let\@oddfoot\@empty
  \let\@evenfoot\@empty}
\makeatother

\maketitle

\section{Introduction}\label{sec1}
\begin{figure}[t]
    \centering
    \includegraphics[width=1.0\textwidth]{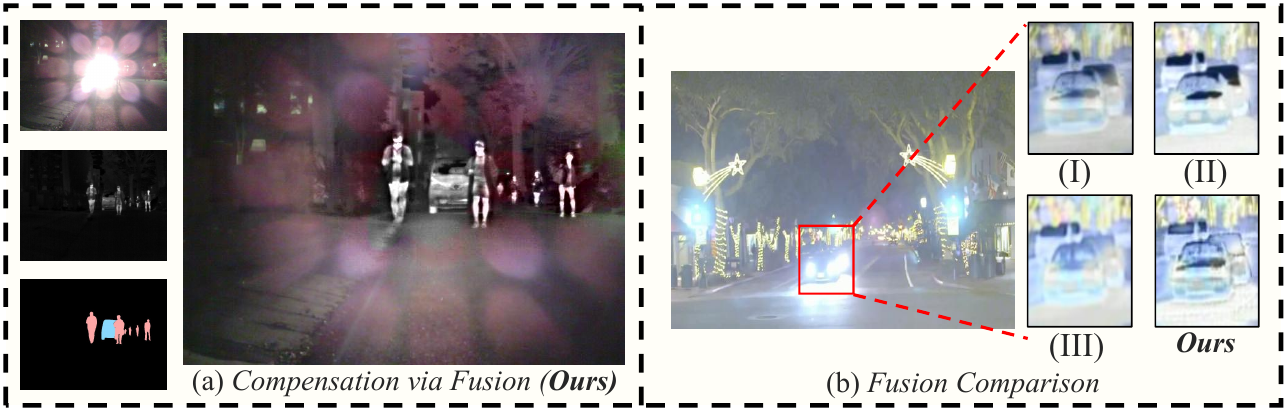}
    \caption{\textcolor{red}{Motivation and visual comparison under overexposed conditions.
    (a) RGB-T fusion compensates for information loss caused by visible overexposure.}
    (b) Local fusion comparison among (I) generative prior-based fusion, (II) downstream task-guided fusion, (III) explicit loss-guided fusion, and Ours.}
    \label{Teaser}
    \vspace{-0.3cm}
\end{figure}

Infrared and visible image fusion plays a key role in applications such as intelligent surveillance~\cite{Jia2021LLVIPAV}  and autonomous driving~\cite{ Liu2023MultiinteractiveFL}. Visible images provide rich texture and structural details, while infrared images capture thermal radiation and offer complementary information for salient targets that is less dependent on variations in illumination. The complementary characteristics of these two modalities enable fused images to provide more comprehensive scene representations~\cite{Tang2023DeepLI}.
In real-world scenarios, strong illumination may saturate visible sensors, causing highlighted regions to lose reliable gradients, textures, and structural cues~\cite{Ceccarelli2020RGBCF, Xie2024OverexposedIA, Qiang2026DWSFusionDW}. In such cases, the visible modality becomes less informative, whereas infrared images can still preserve structural cues of salient objects. \textcolor{red}{As shown in Fig.~\ref{Teaser}(a), image fusion can exploit such complementary cues to compensate for information lost to visible overexposure.} However, most existing fusion methods still struggle with overexposed regions, making robust integration of complementary infrared information under severe overexposure an open challenge.

From the perspective of fusion objective sources, current infrared and visible image fusion methods can be broadly categorized into three groups: generative prior-based~\cite{Ma2019FusionGANAG, Zhao2023DDFMDD,Yue2023DifFusionTH}, downstream task-driven methods~\cite{Liu2022TargetawareDA, Li2024DualModalPS, Chen2025CrosslevelIF} and explicit loss-guided methods~\cite{Zhang2021SDNetAV, Tang2024CAMFAI, Mei2026LearningTO}.
Generative prior-based methods leverage the powerful image generation capability of generative models, such as generative adversarial networks~\cite{goodfellow2014generative} and diffusion models~\cite{ho2020denoising}, to implicitly learn the probability distribution of source images, producing fused results with natural visual fidelity.
Downstream task-driven methods jointly optimize fusion with high-level vision tasks such as detection and segmentation. Semantic information is perceived by the fusion network, and results that better serve downstream tasks are generated beyond visual quality alone.
Explicit loss-guided methods directly constrain the fusion output through carefully designed loss functions. The network is guided to balance visible texture and infrared intensity, and simplicity and computational efficiency can be effectively maintained. However, \textcolor{red}{existing methods still lack fine-grained modeling of local cross-modal reliability under overexposure.}

Recent studies have highlighted the degradation effects of overexposed regions in image fusion~\cite{Ceccarelli2020RGBCF, Xie2024OverexposedIA}. As shown in~\autoref{Teaser}(b), generative prior-based methods prioritize global fidelity over semantic constraints. Although infrared information of the vehicle is preserved in the overexposed region, background noise from the infrared modality is also introduced. Task-driven methods preserve salient structures but often suffer from global degradation, resulting in reduced visual quality. Explicit loss-guided methods are constrained by fixed loss functions that cannot effectively capture semantic complementarity between modalities, leading to either intensity-dominated fusion with information loss or texture-dominated fusion that retains only edges. \textcolor{red}{Therefore, the key challenge is not only to identify where overexposure occurs, but also to determine which degraded regions can benefit from infrared compensation and which infrared structures are reliable for compensation.}

To address the above limitations, we propose EPOFusion, an exposure-aware fusion framework that combines explicit regional guidance with progressive feature refinement. A spatial guidance module identifies infrared compensation regions, while a region-aware fusion loss strengthens informative \textcolor{red}{infrared structures within them}. An iterative feature refinement decoding head with a multiscale context fusion module progressively refines fused representations. It enhances cross-modal information integration, preserves informative infrared structures in saturated regions, and preserves visual consistency in normally exposed areas. 
As shown in~\autoref{Teaser}(b), EPOFusion effectively balances infrared compensation and visible appearance preservation.
Existing public datasets contain limited real-world overexposed samples and \textcolor{red}{generally lack downstream perception annotations,} while existing overexposure-oriented benchmarks provide no explicit guidance for reliable infrared compensation. To fill this gap, we construct the IVOE dataset, including a synthetic training subset with infrared-guided annotations and a real-world subset \textcolor{red}{for evaluating fusion and downstream perception under authentic overexposure.} Extensive experiments demonstrate the competitive fusion and downstream task performance of EPOFusion under challenging overexposure conditions.
The main contributions are summarized as follows:
\begin{itemize}
\item We propose EPOFusion, an exposure-aware fusion framework that leverages a spatial guidance module to \textcolor{red}{identify} infrared \textcolor{red}{compensation regions} and a region-aware loss to strengthen informative infrared structures, effectively preserving infrared details while maintaining visual consistency.

\item We design an iterative feature refinement head with a multiscale context fusion module that progressively refines fused representations, improving detail preservation and robustness under challenging lighting conditions.

\item We build IVOE, \textcolor{red}{which contains 447 real-world image pairs with authentic overexposure and detection and segmentation annotations, together with a synthetic training subset with infrared compensation annotations for guidance supervision.}

\end{itemize}
The remainder of this paper is organized as follows: Section \ref{sec2} overviews the related work. Section \ref{sec3} analyzes fusion challenges under overexposure and presents EPOFusion and the IVOE dataset. Section \ref{sec4} shows the experimental results and analysis, demonstrating the advantages of our method. Section \ref{sec5} concludes the paper.

\section{Related Works}\label{sec2}
\subsection{Generative Prior-based Image Fusion}
Generative prior-based image fusion methods leverage the powerful distribution modeling capabilities of generative models to reconstruct high-quality fused images. Existing methods in this category are broadly categorized into GAN-based~\cite{Ma2019FusionGANAG, Ma2020InfraredAV, Ma2021GANMcCAG} and Denoising Diffusion Probabilistic Model (DDPM)-based~\cite{Yue2023DifFusionTH, Zhao2023DDFMDD, ditfuse2025, Luo2026CUDiffCA}.
FusionGAN~\cite{Ma2019FusionGANAG} pioneered the approach by formulating fusion as a generator–adversarial discriminator problem. Subsequent works like GANMcC~\cite{Ma2021GANMcCAG} enhanced visual fidelity by employing multi-class estimation and incorporating fidelity losses~\cite{Ma2020InfraredAV} to optimize fusion outcomes.
Unlike GAN~\cite{goodfellow2014generative}, which suffers from training instability and mode collapse, DDPM~\cite{ho2020denoising} achieves stable target distribution convergence through a multi-step iterative process. Dif-Fusion~\cite{Yue2023DifFusionTH} first introduced diffusion model, generating high-color-fidelity fused images by constructing the multi-channel input distribution in a latent space. DDFM~\cite{Zhao2023DDFMDD} regards fusion as a conditional generation task, decomposing it into unconditional generation and likelihood rectification based on a hierarchical Bayesian model to achieve fusion without fine-tuning.
Although these methods retain infrared and visible information effectively, they tend to introduce excessive infrared information when handling overexposed scenarios.

 \subsection{Explicit Loss-guided Image Fusion} 
Explicit loss-guided image fusion methods design customized loss functions to guide the model in feature extraction, feature fusion, and pixel-wise reconstruction of the fused image, thereby ensuring high-quality fusion results~\cite{Zhang2021SDNetAV, Tang2024CAMFAI,Bai2024TaskdrivenIF}. SDNet~\cite{Zhang2021SDNetAV} introduces a gradient loss with adaptive decision blocks in pixels, an intensity loss with task-adaptive weighting, and a loss in consistency of decomposition for joint optimization, making it effective and general in diverse fusion tasks. In addition, many existing methods commonly adopt texture loss to preserve structural and detail information~\cite{Dong2024CoEnhancementOM}, along with intensity loss to maintain global brightness and contrast, jointly providing complementary constraints that balance detail preservation and visual fidelity in the fused image. Xie et al.~\cite{Xie2024OverexposedIA} \textcolor{red}{introduce an overexposure prior to identify regions where fusion should rely more on infrared information. However, such region-level guidance does not explicitly distinguish which infrared structures are truly beneficial for compensation. More generally, explicit loss-guided methods struggle to balance infrared compensation and visual consistency under severe overexposure.}

\subsection{Downstream Task-guided Image Fusion}
Image fusion is often used as a preprocessing step for downstream vision tasks. Jointly optimizing image fusion with these tasks has become a growing trend in recent years~\cite{Liu2022TargetawareDA, Liu2023MultiinteractiveFL, Wang2025DiFusionSegDS}. TarDal~\cite{Liu2022TargetawareDA} cascades image fusion with object detection, leveraging the loss of detection to guide the fusion model to preserve more relevant semantic information for detection. SegMif~\cite{Liu2023MultiinteractiveFL} employs a hierarchical interactive attention mechanism within a unified framework to enable bidirectional feature communication between fusion and segmentation tasks. DiFusionSeg~\cite{Wang2025DiFusionSegDS} jointly optimizes fusion and segmentation tasks, achieving \textcolor{red}{competitive} performance in both fusion quality and segmentation. 
However, task-driven methods lack exposure awareness, leading to redundant infrared content in overexposed regions.

\subsection{Infrared and Visible Image Datasets} 
Numerous infrared-visible datasets have been established to facilitate research in infrared and visible image fusion. KAIST~\cite{Hwang2015MultispectralPD}, FMB~\cite{Liu2023MultiinteractiveFL}, MSRS~\cite{Tang2022ImageFI} and M$^3$FD~\cite{Liu2022TargetawareDA} provide abundant aligned image pairs oriented toward driving, with semantic or detection annotations covering typical daytime and nighttime driving scenarios. Xie et al.~\cite{Xie2024OverexposedIA} introduced a dataset focused on overexposure, cropping collected images into $120 \times 120$ patches to construct the training set and selecting 40 pairs for testing. \textcolor{red}{They use an adaptive HSV saturation threshold to detect overexposure and Gaussian fitting to generate a soft prior map. However, existing datasets still lack explicit guidance on reliable infrared structures for compensation.} 
\begin{figure*}[t]
    \centering
    \includegraphics[width=1.0\textwidth]{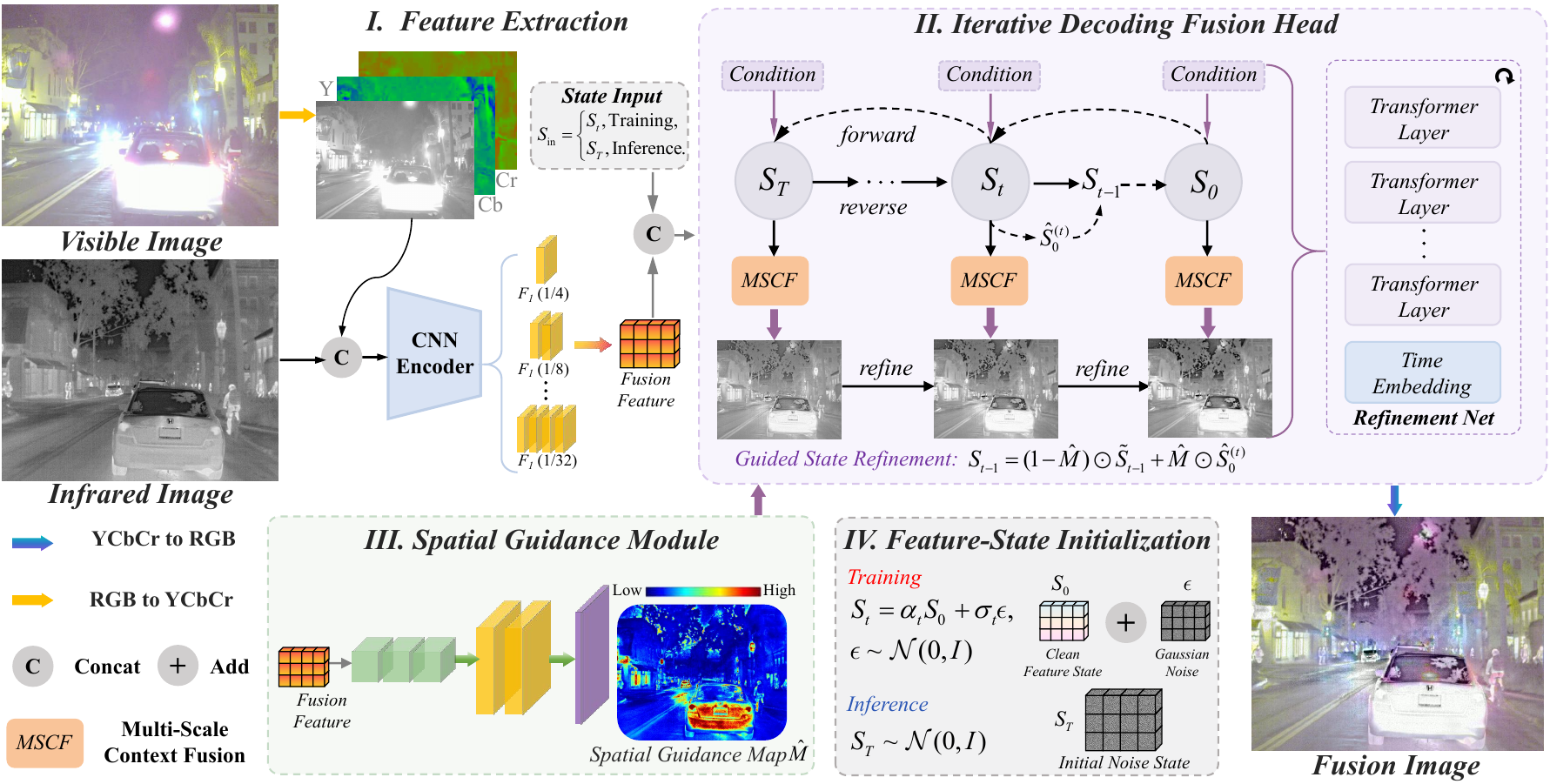}
    \caption{\textcolor{red}{The overall framework of EPOFusion, showing its main components and processing flow.}}
    \label{ModelArch}
    \vspace{-0.6cm}
\end{figure*}

\section{Methodology}\label{sec3}
In this section, we first analyze the problem of detail loss in RGBT image fusion under overexposed conditions. The following section details the proposed EPOFusion framework, along with the construction of the IVOE dataset. 
\subsection{Problem Formulation}
Consider a visible image \(I_{vi}\) and infrared image \(I_{ir}\). The fusion process can be formulated as \(I_f=\mathcal{F}(I_{vi},I_{ir})\), where \(I_f\) denotes fused image. Ideally, \(I_f\) should retain complementary information from both modalities, preserving the structural and textural details of \(I_{vi}\) while incorporating prominent infrared features from \(I_{ir}\). \(I_{vi}\) and \(I_{ir}\) are expected to be recoverable from \(I_f\), which can be formally expressed as follows:
\begin{equation}
    I_{vi}={D}_{vi}(I_f), I_{ir}={D}_{ir}(I_{f}),
\end{equation}
where \({D}_{vi}\) and \({D}_{ir}\) represent modality-specific decoders. 

Explicit loss-guided frameworks typically employ intensity loss \(\mathcal{L}_{int}\) and texture losses \(\mathcal{L}_{tex}\) to guide the fusion process. However, they encounter a critical optimization conflict in overexposed areas. When \(\mathcal{L}_{int}\) dominates, the fused result tends to converge towards the visible image, retaining the saturated intensity in overexposed regions. In contrast, when \(\mathcal{L}_{tex}\) prevails, the fused image inherits the infrared texture but sacrifices crucial thermal intensity information. This dilemma is formulated as follows:
\begin{equation}
    \begin{aligned}
    & I_f \;\Leftarrow\;
     \mathcal{L}_{total} = \lambda_{int}\mathcal{L}_{int} + \lambda_{tex}\mathcal{L}_{tex}, \\[6pt]
    & \rho_{int} =
      \frac{\|\lambda_{int}\nabla_{\theta}L_{int}\|_{2}}
           {\|\nabla_{\theta}L_{total}\|_{2}}, \qquad
      \rho_{tex} =
      \frac{\|\lambda_{tex}\nabla_{\theta}L_{tex}\|_{2}}
           {\|\nabla_{\theta}L_{total}\|_{2}}, \\[8pt]
    & \implies
    \begin{cases}
       I_f^{OE} \approx I_{vi}^{OE}, & \rho_{int} > \rho_{tex}, \\[4pt]
       \| \Delta I_{f}^{OE} - \Delta I_{ir}^{OE} \| < \tau,\; I_f^{OE} \not\approx I_{ir}^{OE}, & \rho_{tex} > \rho_{int},
    \end{cases}
    \end{aligned}
\end{equation}
where \(\lambda_{int}\), \(\lambda_{tex}\) are the weights, $\left\| \cdot \right\|_{2}$  denotes the $\ell_2$  norm, \(\nabla_{\theta}\) represents the gradient with respect to the network parameters, \(OE\) refers to the overexposed area, \(\Delta\) is the Sobel operator, and \(\tau\) denotes an infinitesimal positive value, indicating that the two gradient responses are nearly identical.

Downstream task-driven fusion methods that prioritize texture are prone to conflicts in overexposed scenes. Specifically, the smooth, featureless regions of the visible image are juxtaposed with the low-signal infrared counterparts, where sensor noise predominates. Similarly, generative prior--based fusion models struggle to discriminate between informative content and spurious artifacts, often retaining all signals indiscriminately. Due to these shortcomings, the model fails to distinguish meaningful texture from noise, misinterprets the noise as a salient feature, and erroneously introduces it into the final image.

The goal of this paper is twofold: (i) \textcolor{red}{to identify regions where infrared information provides valid compensation for visible degradation; and} (ii) \textcolor{red}{to effectively preserve such complementary information while suppressing redundant infrared content.}

\subsection{Exposure Guided Fusion Framework}
Overexposure often leads to the loss of critical visible details. To address this issue, we propose an exposure-aware fusion framework, as illustrated in~\autoref{ModelArch}. The proposed framework aims to \textcolor{red}{identify locally reliable infrared information for regions degraded by overexposure and progressively incorporate it into the fused representation}. Specifically, the framework integrates multimodal feature extraction, spatial region guidance, and an iterative feature refinement decoding head \textcolor{red}{under region-aware fusion supervision to better integrate useful and reliable information from both modalities}.

For each pair of \(I_{vi}\) and \(I_{ir}\), we first convert \(I_{vi}\) from RGB to the YCbCr color space, with only the luminance (Y) channel employed as input. We concatenate the infrared image with the visible Y channel and feed them into a ConvNeXt encoder~\cite{Liu2022ACF} to extract multimodal features \(\mathcal{F}_{fusion}\). Based on \(\mathcal{F}_{fusion}\), the spatial guidance module identifies visible regions degraded by overexposure and determines where reliable infrared information can provide effective compensation. \textcolor{red}{Such region-aware guidance further modulates the iterative state refinement, enabling the decoder to progressively integrate reliable and complementary information into the fused representation}. The decoder then predicts the fused Y channel, which is combined with the original Cb and Cr channels to reconstruct the final RGB image.

As illustrated in~\autoref{ModelArch}, the guidance heatmaps highlight informative infrared structures within degraded visible regions. \textcolor{red}{By coupling such spatial guidance with iterative feature refinement, EPOFusion selectively strengthens useful cross-modal information while suppressing irrelevant infrared content}, thereby preserving fine structures and maintaining robust fusion performance under overexposure conditions.

\begin{figure}[t]
    \centering
    \includegraphics[width=0.60\textwidth]{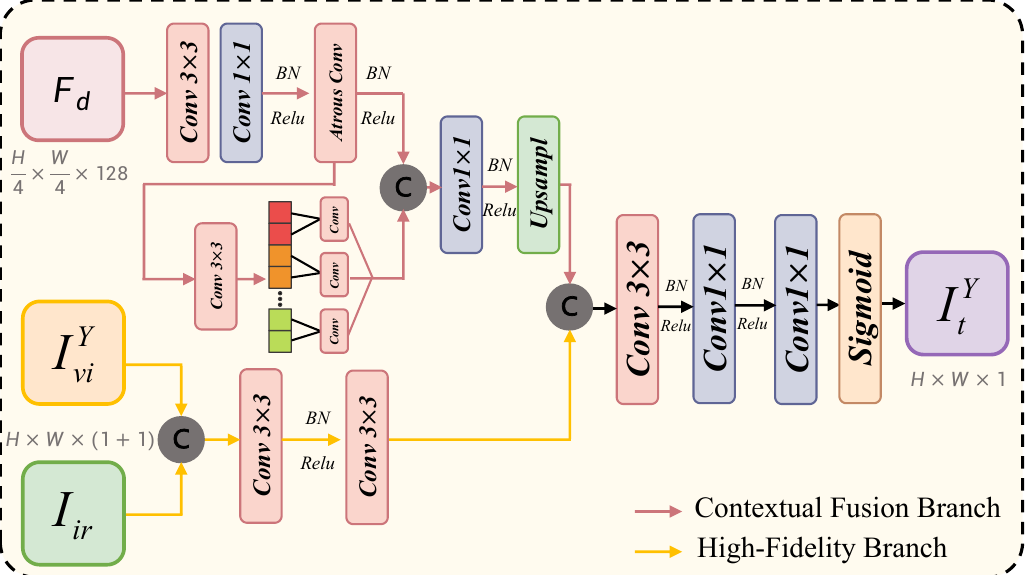}
    \caption{The architecture of the Multi-Scale Context Fusion Module (MSCF).}
    \label{MSCF}
    \vspace{-0.6cm}
\end{figure}
\subsection{Iterative Feature Refinement Decoding Head}
Traditional fusion decoders transform the fused feature into image space in a single step. Nevertheless, in challenging scenarios such as high-brightness regions, this one-shot approach often fails to fully exploit the fused feature, leading to incomplete multimodal information integration. To overcome this limitation, we design an iterative decoding strategy that \textcolor{red}{progressively refines the multimodal feature and reconstructs intermediate fused outputs}, enabling complementary cross-modal information to be gradually integrated.

\begin{figure*}[htbp]
\vspace{-0.3cm}
\centering
\begin{minipage}[t]{0.45\textwidth}
    \begin{algorithm}[H]
        \caption{: Training}\label{algorithm_train}
        \begin{algorithmic}
        \setlength{\baselineskip}{1.32em}
        \STATE
        \STATE \textbf{def} train(\(I_{vi}, I_{ir}, \textcolor{red}{M, I_{vi}^{clean}}\)) :
        \STATE \hspace{0.40cm}\(F_{fusion}=Encoder(I_{vi},I_{ir})\) 
        \STATE \hspace{0.40cm}\(\hat{M} = \mathrm{Guide}(F_{fusion})\)
        \STATE \hspace{0.40cm}\textcolor{red}{\(S_0=Norm(Encoder(I_{vi}^{clean},I_{ir}))\)} 
        \STATE \hspace{0.40cm}\textcolor{black}{\textbf{for} \(t\in\{t_1,\ldots,t_T\}\) \textbf{do}}
        \STATE \hspace{0.78cm}\(\epsilon_t\sim\mathcal{N}(0,I)\)
        \STATE \hspace{0.78cm}\textcolor{red}{\(S_t=\alpha_tS_0+\sigma_t\epsilon_t\)}
        \STATE \hspace{0.78cm}\textcolor{red}{\(F_{cond}=F_{fusion} \oplus S_t\)}
        \STATE \hspace{0.78cm}\(F_d=D_{\theta}(\textcolor{red}{F_{cond}},t)\)
        \STATE \hspace{0.78cm}\textcolor{red}{\(\hat{S}_0^{(t)}=Norm(F_d)\)}
        \STATE \hspace{0.78cm}\textcolor{black}{\(I_t\)}\(=MSCF(F_d,I_{vi},I_{ir})\)
        \STATE \textcolor{red}{\textbf{return} $\mathcal{L}$ from Eq.~(10)}
        \end{algorithmic}
    \end{algorithm}
\end{minipage}
\hfill
\begin{minipage}[t]{0.45\textwidth}
    \begin{algorithm}[H]
        \caption{: Inference}\label{algorithm_inference}
        \begin{algorithmic}
        \setlength{\baselineskip}{1.32em}
        \STATE
        \STATE \textbf{def} inference(\(I_{vi},I_{ir}\)) :
        \STATE \hspace{0.40cm} \(F_{fusion}=Encoder(I_{vi},I_{ir})\)
        \STATE \hspace{0.40cm} \textcolor{red}{\(\hat{M}=Guide(F_{fusion})\)}
        \STATE \hspace{0.40cm} \textcolor{red}{\(S_T\sim\mathcal{N}(0,I)\)}
        \STATE \hspace{0.40cm} \textbf{for} \(t=T,\ldots,1\) \textbf{do}
        \STATE \hspace{0.78cm} \textcolor{red}{\(F_d=D_\theta(F_{fusion}\oplus S_t,t)\)}
        \STATE \hspace{0.78cm} \(I_t=MSCF(F_d,I_{vi},I_{ir})\)
        \STATE \hspace{0.78cm} \textcolor{red}{\(\hat{S}_0^{(t)}=Norm(F_d)\)}
        \STATE \hspace{0.78cm} \textcolor{red}{
        \(\hat{\epsilon}_t=(S_t-\alpha_t\hat{S}_0^{(t)})/\sigma_t\)}
        \STATE \hspace{0.78cm} \textcolor{red}{
        \(\tilde{S}_{t-1}
        =\alpha_{t-1}\hat{S}_0^{(t)}
        +\sigma_{t-1}\hat{\epsilon}_t\)}
        \STATE \hspace{0.78cm} \textcolor{red}{
        \(S_{t-1}
        =(1-\hat{M})\odot\tilde{S}_{t-1}
        +\hat{M}\odot\hat{S}_0^{(t)}\)}
        \hspace{0.40cm} \textbf{return} \(I_t\)
        \end{algorithmic}
    \end{algorithm}
\end{minipage}
\vspace{-0.6cm}
\end{figure*}

\subsubsection{Iterative Refinement Mechanism} 
The overall training and inference processes of EPOFusion are summarized in Algorithms \ref{algorithm_train} and \ref{algorithm_inference}. Specifically, we formulate fusion as a progressive feature refinement process, where latent representations are iteratively refined under exposure-aware guidance. \textcolor{red}{Since overexposure is synthetically introduced during training, the clean visible image remains available. This enables us to construct a reference state from the original visible-infrared pair to provide a feature-space reference without requiring ground-truth fused images.} The proposed process adopts a DDIM-style~\cite{Song2020DenoisingDI} deterministic update for conditional feature refinement under the guidance of $F_{fusion}$.

In the training process, the iterative decoder follows a conditional refinement paradigm, where the fusion feature $F_{fusion}$ serves as the primary conditional input. \textcolor{red}{Specifically, the reference state \(S_0\) is obtained as}: 
\begin{equation} 
\textcolor{red}{ S_0=Norm\left(Encoder(I_{vi}^{clean},I_{ir})\right). } 
\end{equation} 
\textcolor{red}{Based on the reference state \(S_0\), a noisy feature state $S_t$ is independently constructed at each timestep according to the noise schedule:}
\begin{equation}
\textcolor{red}{
S_t}=\alpha_t \textcolor{red}{S_0}+\sigma_t\epsilon,\qquad \epsilon\sim\mathcal{N}(0,I),
\end{equation}
where \(\alpha_t\) and \(\sigma_t\) denote the noise coefficients. 
\textcolor{red}{A cosine log-SNR schedule was adopted to determine \(\alpha_t\) and \(\sigma_t\) at different timesteps.}
\textcolor{red}{The resulting state \(S_t\) is then combined with the multimodal feature \(F_{fusion}\)} and fed into the refinement network \(D_{\theta}\):
\begin{equation}
\textcolor{red}{F_{cond}=F_{fusion}\oplus S_t},\quad F_d=D_{\theta}(F_{cond},t),\quad \textcolor{red}{ \hat{S}_0^{(t)}=Norm(F_d)},
\end{equation}
where \(\oplus\) denotes feature concatenation. \textcolor{red}{The predicted state $\hat{S}_0^{(t)}$ is constrained toward the reference state $S_0$, while $F_d$} is decoded by the MSCF module to generate the corresponding fused output $I_t$.

During inference, we first sample \textcolor{red}{an initial feature state $S_T\sim\mathcal{N}(0,I)$}. The model then refines the state for \textcolor{red}{$T$ steps}. At each step, the current feature state \textcolor{red}{$S_t$} is combined with $F_{fusion}$ and processed by the refinement network $\mathcal{D}_{\theta}$ to obtain \textcolor{red}{$F_d$}, which is normalized to produce \textcolor{red}{$\hat{S}_0^{(t)}$}. The refined feature is further decoded by the MSCF module to obtain the intermediate fused image \textcolor{red}{$I_t$}. The residual is then estimated \textcolor{red}{from the current state and clean-state proposal} and used to update the feature state for the next step. \textcolor{red}{The predicted compensation map $\hat{M}$ further modulates this transition to strengthen refinement in infrared compensation regions.} The process can be expressed as:
\begin{equation}
    \begin{aligned}
    &(\alpha_t,\sigma_t,\alpha_{t-1},\sigma_{t-1})
    =Cal(t_{now},t_{next}),\\
    &\textcolor{red}{\hat{\epsilon}t=
    \frac{S_t-\alpha_t\hat{S}_0^{(t)}}{\sigma_t}},\\
    &\textcolor{red}{\tilde{S}_{t-1}
    =\alpha_{t-1}\hat{S}_0^{(t)}
    +\sigma_{t-1}\hat{\epsilon}_t},\\
    &\textcolor{red}{S_{t-1}
    =(1-\hat{M})\odot\tilde{S}_{t-1}
    +\hat{M}\odot\hat{S}_0^{(t)}},
    \end{aligned}
\end{equation}
where $Cal(\cdot)$ computes the DDIM coefficients between two consecutive timesteps, and \textcolor{red}{$\tilde{S}_{t-1}$ denotes the DDIM-updated feature before compensation-guided modulation}. \textcolor{red}{At each refinement step, $\hat{M}$ increases the contribution of $\hat{S}_0^{(t)}$ in compensation regions, promoting the progressive integration of informative infrared structures.}

\subsubsection{Multi-Scale Context Fusion Module}
To effectively leverage the refined features from the iterative optimizer and enhance multi-scale, cross-modal feature interactions, we introduce the MSCF module. The MSCF is composed of two specialized pathways: a Contextual Fusion Branch (CFB) and a High-Fidelity Branch (HFB), designed to reconstruct high-quality fused images.

The CFB, as illustrated in~\autoref{MSCF}, is designed to extract multi-scale contextual information from the refined feature $F_{d}$. To reduce computational redundancy, the input is first processed by sequential convolutions. Subsequently, to capture both long-range dependencies and local details, we employ a dual-branch architecture. The first branch utilizes dilated convolutions to expand the receptive field, thereby gathering global contextual information. The second branch employs a cascade of standard and grouped convolutions to focus on fine-grained local features. Finally, the features from both branches are concatenated and fused via a $1\times1$ convolution to generate the high-level context representation $F_{CFB}$ This process is formulated as:
\begin{equation}
    \begin{aligned}
        & F_{global} = \sigma_{R}({BN}({Conv}_{{Atrous}}({Conv}(F_{d})))), \\
        & F_{s} = {Conv}_s(({F}_{d})_s), \quad s=1,2,\ldots,G, \\
        & F_{fused} = {Conv}_{1\times1}(F_{global} \oplus (F_1 \oplus F_2 \cdots \oplus F_G)), \\
        & F_{CFB} = \sigma_{R}({BN}(F_{fused})),
    \end{aligned}
\end{equation}
where \(G\) denotes the total number of groups, \(BN(\cdot)\) represents Batch Normalization, \(Conv_{Atrous}(\cdot)\) indicates the atrous convolution, \(Conv_{1\times 1}(\cdot)\) refers to the point-wise convolution, and \(\sigma_R(\cdot)\) is the relu activation function.

The HFB is designed to preserve fine-grained details from the source images. It begins by concatenating the visible image \(I_{vi}^Y\) and the infrared image \(I_{ir}\) to create a dual-channel input. This is then processed by two consecutive convolution layers that extract essential low-level features, such as edges and textures, producing a detailed feature map \(F_{HFB}\). This operation is formulated as:
\begin{equation}
    F_{HFB}=\sigma_R\Big(BN({Conv}({Conv}(I_{vi}^{Y} \oplus I_{ir})))\Big).
\end{equation}

Finally, the features from both branches are fused. The high-level context from \(F_{\text{CFB}}\) and the detailed information from \(F_{\text{HFB}}\) are concatenated and passed through a refinement block \(\Phi\). This allows the network to intelligently balance and select complementary information from both sources. A final Sigmoid activation function is then applied to generate the fused luminance output\(I_t^Y\) at refinement step $t$. The overall process can be expressed as:
\begin{equation}
    \begin{aligned}
        &I_t^{Y}=\sigma_{S}({\Phi(F_{CFB} \oplus F_{HFB})}), 
    \end{aligned}
\end{equation}
where \(\Phi(\cdot)\) denotes convolutional nonlinear mapping, and $\sigma_S(\cdot)$ is the sigmoid function.

\begin{figure}[t]
    \centering
    \includegraphics[width=1.0\textwidth]{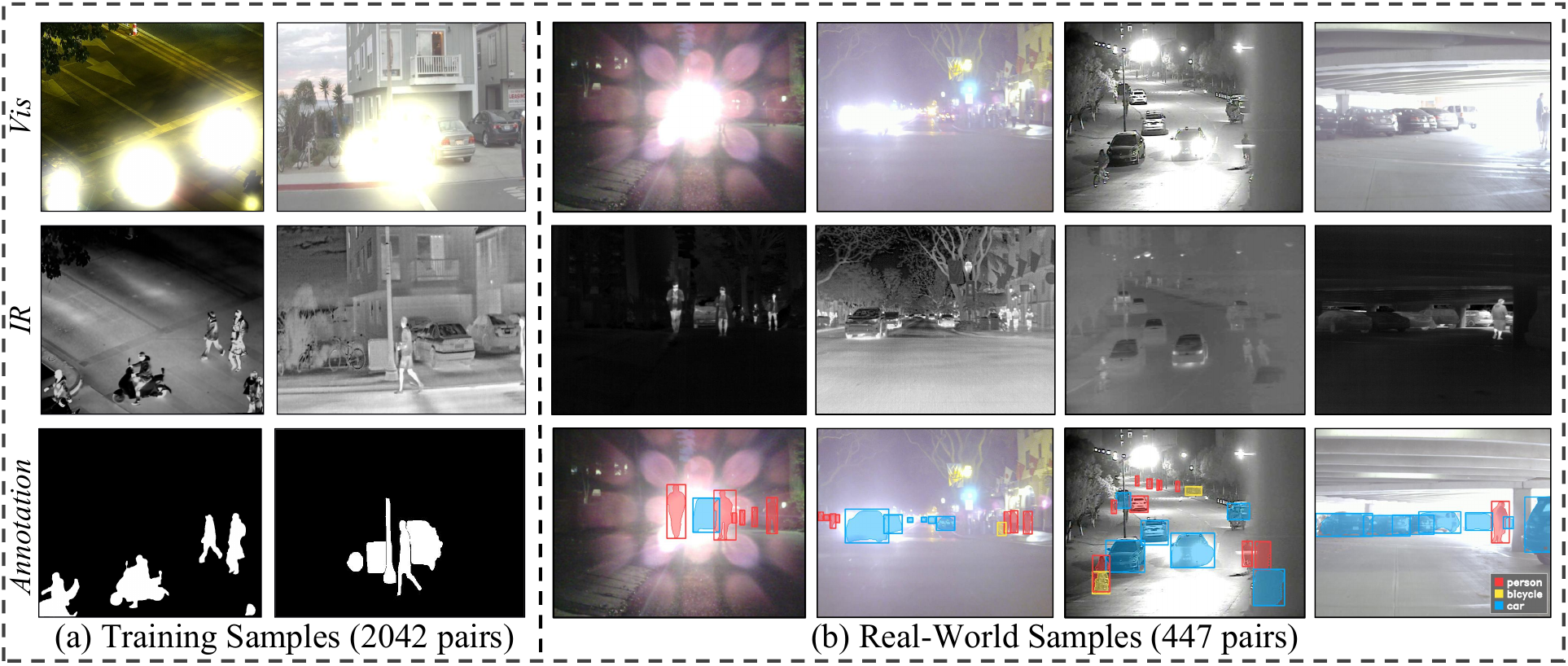}
    \caption{Examples from the IVOE dataset, \textcolor{red}{including a synthetic training subset with infrared compensation masks and a real-world subset of 447 pairs annotated for detection and segmentation.}}
    \label{Dataset}
    \vspace{-0.6cm}
\end{figure}

\subsection{Infrared-Visible Overexposure Dataset}\label{IVOE}
Existing public datasets provide limited overexposed samples and generally lack \textcolor{red}{supervision for identifying reliable infrared information in degraded regions}. To address this, we construct IVOE with a synthetic training subset \textcolor{red}{for controllable exposure-compensation learning} and a real-world subset for generalization under authentic overexposure conditions.

\subsubsection{Synthetic Training Set}
The synthetic training set contains 2,042 aligned infrared-visible image pairs from the training splits of MSRS~\cite{Tang2022ImageFI}, FMB~\cite{Liu2023MultiinteractiveFL}, LLVIP~\cite{Jia2021LLVIPAV}, and FLIR-ADAS\footnote{\url{https://www.flir.com/oem/adas/adas-dataset-form/}}, \textcolor{red}{including 935, 676, 230, and 201 pairs, respectively.}
\textcolor{red}{We apply target-guided Gaussian exposure perturbations with varying scales, intensities, and colors to salient objects in visible images, producing controllable local degradation. The original visible images are retained as aligned photometric references for the degraded regions.} SAM~\cite{Kirillov2023SegmentA} is used to assist candidate-region annotation, followed by manual verification and correction of the masks to ensure that the selected regions contain reliable infrared structures for compensation. Representative examples are shown in~\autoref{Dataset}(a). \textcolor{red}{Unlike conventional exposure masks that only indicate degraded visible regions, our masks identify regions where visible information is degraded while infrared retains reliable structures, thereby providing explicit infrared compensation guidance.}

\subsubsection{Real-World Subset}
To evaluate generalization under authentic overexposure conditions, we construct a real-world subset containing \textcolor{red}{447} infrared-visible image pairs. The set includes \textcolor{red}{142} pairs from MSRS~\cite{Tang2022ImageFI}, \textcolor{red}{8} from LLVIP~\cite{Jia2021LLVIPAV}, \textcolor{red}{15} from FMB~\cite{Liu2023MultiinteractiveFL}, \textcolor{red}{5} from VIFB~\cite{Zhang2020VIFBAV}, \textcolor{red}{and 277 from FLIR-ADAS.} 
\textcolor{red}{It covers diverse overexposure scenarios, including daytime street scenes under strong illumination and nighttime driving scenes with local overexposure caused by headlights or other strong light sources. Candidate samples are screened using a core ratio $\geq 25\%$, halo ratio $\geq 45\%$, RGB gradient $\leq 6$, IR--RGB gradient gain $\geq 0.6$, and dynamic-range gain $\geq 12$, followed by manual review. This protocol targets scenes where infrared information provides effective compensation.} 
Representative examples are shown in~\autoref{Dataset}(b).

\textcolor{red}{Since the visible and infrared images in FLIR-ADAS are captured by separate cameras and are not pixel-wise aligned, we perform cross-modal dense registration and crop the common valid regions to obtain aligned $640 \times 512$ image pairs. The real-world subset is further annotated for object detection and semantic segmentation, covering person, car, and bicycle. It contains 3,823 bounding boxes, including 1,994 persons, 1,640 cars, and 189 bicycles, and pixel-level semantic masks for all 447 image pairs. Bounding boxes are manually annotated, while masks are SAM-assisted and manually refined.}  All real-world samples are excluded from EPOFusion training and parameter tuning. They are used for fusion evaluation and further split for downstream model training and testing, while the fusion model remains fixed.

\subsection{Loss Function}
EPOFusion is jointly optimized by a fusion loss, \textcolor{red}{a feature-state loss}, and a guidance loss, which respectively constrain the fused output, iterative feature refinement, and infrared compensation regions.
\begin{equation}
\mathcal{L}=\textcolor{red}{\frac{1}{T}\sum_{t=1}^{T}\left(\mathcal{L}_{fusion}^{(t)}
+
\lambda_s\mathcal{L}_{state}^{(t)}\right)}+\lambda_g\mathcal{L}_{guidance},
\end{equation}
\textcolor{red}{where $T$ denotes the number of refinement time steps, and $\lambda_s$ and $\lambda_g$ are set to 0.05 and 0.4, respectively. The feature-state loss is defined as:}
\begin{equation}
    \textcolor{red}{
    \mathcal{L}_{state}^{(t)}=\operatorname{MSE}\left(\hat{S}_0^{(t)},S_0\right).
    }
\end{equation}

The image is \textcolor{red}{softly partitioned by a compensation weight $g\in[0,1]$ derived from $M$, with $w_N=1-g$. $g$ indicates regions requiring infrared compensation, while $w_N$ represents the remaining regions. Here, $I_{vi}$ denotes the degraded visible input to the fusion network.} For regions weighted by \textcolor{red}{$w_N$}, $\mathcal{L}_{in}$ preserves the stronger modal response.
\begin{equation}
    \mathcal{L}_\mathrm{in}^\mathrm{N}=\left\|\textcolor{red}{w_N}\odot\left(\textcolor{red}{I_t}-\max(I_{vi},I_{ir})\right)\right\|_1,
\end{equation}
where \(\odot\) denotes element-wise product, \(\| \cdot \|_1\) denotes the $\ell_1$  norm. \textcolor{red}{For regions weighted by $g$, an infrared-enhanced target is defined as:}
\begin{equation}
    \textcolor{red}{
    \begin{aligned}
    T_G &= \operatorname{LP}\left(\max(I_{vi},I_{ir})\right)
    +\gamma\operatorname{HP}(I_{ir}),\\
    \mathcal{L}_\mathrm{in}^{G}
    &=\left\|g\odot\left(I_t-T_G\right)\right\|_1 ,
    \end{aligned}
    }
\end{equation}
\textcolor{red}{where $\operatorname{LP}(\cdot)$ denotes the low-frequency component and $\operatorname{HP}(I)=I-\operatorname{LP}(I)$ denotes the corresponding high-frequency component.} To preserve structural details, a gradient loss \(\mathcal{L}_{grad}\) is also applied. In the regions weighted by $w_N$, it preserves prominent gradients from both modalities:
\begin{equation}
    \mathcal{L}_{\mathrm{grad}}^{\textcolor{red}{\mathrm{N}}}=\left\|\textcolor{red}{w_N}\odot\left(\Delta I_t-\max(\Delta I_{vi},\Delta I_{ir})\right)\right\|_1.
\end{equation} 

In the regions weighted by \textcolor{red}{$g$}, the gradient loss emphasizes reliable and informative infrared structures:
\begin{equation}
    \mathcal{L}_\mathrm{grad}^{G}=\left\|\textcolor{red}{g}\odot\left(\nabla  \textcolor{red}{I_t}-\gamma\nabla I_{ir}\right)\right\|_1,
\end{equation}
\textcolor{red}{where $\nabla$ denotes the signed Sobel gradient and $\gamma=2.5$.}  The total fusion loss \textcolor{red}{at each step} is given by:
\begin{equation}
    \textcolor{red}{
    \mathcal{L}_\mathrm{fusion}^{(t)}=\sum_{r\in\{N,G\}}\left(\mathcal{L}_\mathrm{in}^{r,(t)}
    +
    \mathcal{L}_\mathrm{grad}^{r,(t)}\right).
    }
\end{equation}

The spatial guidance module utilizes the cross-entropy loss function, defined as: 
\begin{equation} \textcolor{red}{ \mathcal{L}_{guidance} =-\frac{1}{n}\sum_{i=1}^{n}\sum_{c=0}^{1} q_{i,c}\log p_{i,c}}, \label{seg_main} 
\end{equation} 
\textcolor{red}{where $q_{i,c}$ and $p_{i,c}$ denote the ground-truth indicator and predicted probability for class $c$, respectively, and $n$ is the number of valid pixels.} 

Collectively, these loss components enable EPOFusion to effectively handle overexposed regions while preserving both structural fidelity and visual consistency.
\begin{table}[t!]
    \centering
    \caption{Quantitative comparisons on 361 image pairs from the MSRS dataset. 
    The \colorbox{bestcell}{\textbf{red}} and \colorbox{secondcell}{blue} marks indicate the best and second-best values, respectively. 
    Mask-D. represents Mask-DiFuser.}
    \label{MSRS}
    \fontsize{10pt}{12pt}\selectfont
    \setlength{\tabcolsep}{3pt}
    \renewcommand{\arraystretch}{1.1}
    \begin{tabular*}{\columnwidth}{@{\extracolsep{\fill}}l*{6}{>{\centering\arraybackslash}c}}
        \toprule
        \textbf{Category} 
        & MI$\uparrow$ 
        & VIF$\uparrow$ 
        & \(Q^{AB/F}\)$\uparrow$ 
        & SSIM$\uparrow$ 
        & MS\_SSIM$\uparrow$ 
        & PI$\downarrow$ \\
        \midrule

        FusionGAN~\cite{Ma2019FusionGANAG}
        & ${1.692}_{\scriptscriptstyle \pm 0.153}$
        & ${0.422}_{\scriptscriptstyle \pm 0.038}$
        & ${0.235}_{\scriptscriptstyle \pm 0.065}$
        & ${0.281}_{\scriptscriptstyle \pm 0.032}$
        & ${0.404}_{\scriptscriptstyle \pm 0.011}$
        & ${4.198}_{\scriptscriptstyle \pm 0.169}$ \\

        U2Fusion~\cite{Xu2020U2FusionAU}
        & ${2.452}_{\scriptscriptstyle \pm 0.402}$
        & ${0.666}_{\scriptscriptstyle \pm 0.130}$
        & ${0.472}_{\scriptscriptstyle \pm 0.075}$
        & ${0.441}_{\scriptscriptstyle \pm 0.052}$
        & ${0.497}_{\scriptscriptstyle \pm 0.050}$
        & ${4.045}_{\scriptscriptstyle \pm 0.007}$ \\

        GANMcC~\cite{Ma2021GANMcCAG}
        & ${2.306}_{\scriptscriptstyle \pm 0.133}$
        & ${0.595}_{\scriptscriptstyle \pm 0.019}$
        & ${0.286}_{\scriptscriptstyle \pm 0.041}$
        & ${0.369}_{\scriptscriptstyle \pm 0.052}$
        & ${0.463}_{\scriptscriptstyle \pm 0.018}$
        & ${4.627}_{\scriptscriptstyle \pm 0.119}$ \\

        SDNet~\cite{Zhang2021SDNetAV}
        & ${1.939}_{\scriptscriptstyle \pm 0.005}$
        & ${0.509}_{\scriptscriptstyle \pm 0.001}$
        & ${0.400}_{\scriptscriptstyle \pm 0.002}$
        & ${0.399}_{\scriptscriptstyle \pm 0.000}$
        & ${0.452}_{\scriptscriptstyle \pm 0.000}$
        & ${4.034}_{\scriptscriptstyle \pm 0.004}$ \\

        TarDal~\cite{Liu2022TargetawareDA}
        & ${2.818}_{\scriptscriptstyle \pm 0.021}$
        & ${0.605}_{\scriptscriptstyle \pm 0.002}$
        & ${0.370}_{\scriptscriptstyle \pm 0.003}$
        & ${0.427}_{\scriptscriptstyle \pm 0.012}$
        & ${0.484}_{\scriptscriptstyle \pm 0.004}$
        & ${4.279}_{\scriptscriptstyle \pm 0.013}$ \\

        SegMif~\cite{Liu2023MultiinteractiveFL}
        & ${3.072}_{\scriptscriptstyle \pm 0.016}$
        & ${0.819}_{\scriptscriptstyle \pm 0.006}$
        & ${0.596}_{\scriptscriptstyle \pm 0.001}$
        & ${0.436}_{\scriptscriptstyle \pm 0.001}$
        & ${0.476}_{\scriptscriptstyle \pm 0.001}$
        & ${3.683}_{\scriptscriptstyle \pm 0.008}$ \\

        DDFM~\cite{Zhao2023DDFMDD}
        & ${2.615}_{\scriptscriptstyle \pm 0.067}$
        & ${0.704}_{\scriptscriptstyle \pm 0.028}$
        & ${0.471}_{\scriptscriptstyle \pm 0.018}$
        & ${0.383}_{\scriptscriptstyle \pm 0.079}$
        & ${0.489}_{\scriptscriptstyle \pm 0.036}$
        & ${4.118}_{\scriptscriptstyle \pm 0.003}$ \\

        Dif-Fusion~\cite{Yue2023DifFusionTH}
        & ${3.331}_{\scriptscriptstyle \pm 0.006}$
        & ${0.828}_{\scriptscriptstyle \pm 0.001}$
        & ${0.583}_{\scriptscriptstyle \pm 0.000}$
        & ${0.448}_{\scriptscriptstyle \pm 0.000}$
        & ${0.506}_{\scriptscriptstyle \pm 0.000}$
        & ${3.588}_{\scriptscriptstyle \pm 0.010}$ \\

        DCINN~\cite{Wang2023AGP}
        & ${\second{3.581}}_{\scriptscriptstyle \pm 0.011}$
        & ${0.887}_{\scriptscriptstyle \pm 0.001}$
        & ${0.626}_{\scriptscriptstyle \pm 0.000}$
        & ${\second{0.491}}_{\scriptscriptstyle \pm 0.001}$
        & ${0.511}_{\scriptscriptstyle \pm 0.000}$
        & ${4.263}_{\scriptscriptstyle \pm 0.004}$ \\

        MRFS~\cite{Zhang2024MRFSMR}
        & ${2.980}_{\scriptscriptstyle \pm 0.077}$
        & ${0.715}_{\scriptscriptstyle \pm 0.021}$
        & ${0.475}_{\scriptscriptstyle \pm 0.008}$
        & ${0.347}_{\scriptscriptstyle \pm 0.003}$
        & ${0.449}_{\scriptscriptstyle \pm 0.002}$
        & ${5.408}_{\scriptscriptstyle \pm 0.006}$ \\

        CCF~\cite{cao2024conditional}
        & ${2.203}_{\scriptscriptstyle \pm 0.093}$
        & ${0.595}_{\scriptscriptstyle \pm 0.025}$
        & ${0.427}_{\scriptscriptstyle \pm 0.007}$
        & ${0.382}_{\scriptscriptstyle \pm 0.020}$
        & ${0.500}_{\scriptscriptstyle \pm 0.005}$
        & ${3.682}_{\scriptscriptstyle \pm 0.017}$ \\

        A2RNet~\cite{Li2024A2RNetAA}
        & ${3.347}_{\scriptscriptstyle \pm 0.031}$
        & ${0.640}_{\scriptscriptstyle \pm 0.062}$
        & ${0.414}_{\scriptscriptstyle \pm 0.021}$
        & ${0.332}_{\scriptscriptstyle \pm 0.023}$
        & ${0.456}_{\scriptscriptstyle \pm 0.019}$
        & ${6.391}_{\scriptscriptstyle \pm 0.010}$ \\

        DiFusionSeg~\cite{Wang2025DiFusionSegDS}
        & ${3.517}_{\scriptscriptstyle \pm 0.008}$
        & ${0.818}_{\scriptscriptstyle \pm 0.007}$
        & ${\second{0.627}}_{\scriptscriptstyle \pm 0.001}$
        & ${0.447}_{\scriptscriptstyle \pm 0.004}$
        & ${0.511}_{\scriptscriptstyle \pm 0.001}$
        & ${\second{3.387}}_{\scriptscriptstyle \pm 0.060}$ \\

        MaeFuse~\cite{10893688}
        & ${2.133}_{\scriptscriptstyle \pm 0.058}$
        & ${0.747}_{\scriptscriptstyle \pm 0.001}$
        & ${0.501}_{\scriptscriptstyle \pm 0.001}$
        & ${0.435}_{\scriptscriptstyle \pm 0.004}$
        & ${\best{0.525}}_{\scriptscriptstyle \pm 0.001}$
        & ${4.708}_{\scriptscriptstyle \pm 0.023}$ \\

        Mask-D.~\cite{11162636}
        & ${2.391}_{\scriptscriptstyle \pm 0.009}$
        & ${\second{0.912}}_{\scriptscriptstyle \pm 0.005}$
        & ${0.512}_{\scriptscriptstyle \pm 0.002}$
        & ${0.359}_{\scriptscriptstyle \pm 0.011}$
        & ${0.503}_{\scriptscriptstyle \pm 0.002}$
        & ${\best{3.029}}_{\scriptscriptstyle \pm 0.030}$ \\

        \textcolor{red}{OpIVF}~\cite{Xie2024OverexposedIA}
        & ${3.079}_{\scriptscriptstyle \pm 0.044}$
        & ${0.846}_{\scriptscriptstyle \pm 0.001}$
        & ${0.603}_{\scriptscriptstyle \pm 0.002}$
        & ${0.481}_{\scriptscriptstyle \pm 0.001}$
        & ${0.488}_{\scriptscriptstyle \pm 0.000}$
        & ${3.723}_{\scriptscriptstyle \pm 0.020}$ \\

        \textbf{EPOFusion$^{*}$}
        & ${\best{3.615}}_{\scriptscriptstyle \pm 0.010}$
        & ${\best{0.921}}_{\scriptscriptstyle \pm 0.007}$
        & ${\best{0.655}}_{\scriptscriptstyle \pm 0.003}$
        & ${\best{0.495}}_{\scriptscriptstyle \pm 0.002}$
        & ${\second{0.521}}_{\scriptscriptstyle \pm 0.001}$
        & ${3.423}_{\scriptscriptstyle \pm 0.036}$ \\

        \bottomrule
    \end{tabular*}
    \vspace{-0.3cm}
\end{table}
\begin{table}[t!]
    \centering
    \caption{Quantitative comparisons on 280 image pairs from the FMB dataset.}
    \label{FMB}
    \fontsize{10pt}{12pt}\selectfont
    \setlength{\tabcolsep}{3pt}
    \renewcommand{\arraystretch}{1.1}
    \begin{tabular*}{\columnwidth}{@{\extracolsep{\fill}}l*{6}{>{\centering\arraybackslash}c}}
        \toprule
        \textbf{Category} 
        & MI$\uparrow$ 
        & VIF$\uparrow$ 
        & \(Q^{AB/F}\)$\uparrow$ 
        & SSIM$\uparrow$ 
        & MS\_SSIM$\uparrow$ 
        & PI$\downarrow$ \\
        \midrule

        FusionGAN~\cite{Ma2019FusionGANAG}
        & ${3.432}_{\scriptscriptstyle \pm 0.081}$
        & ${0.502}_{\scriptscriptstyle \pm 0.007}$
        & ${0.497}_{\scriptscriptstyle \pm 0.002}$
        & ${0.404}_{\scriptscriptstyle \pm 0.005}$
        & ${0.458}_{\scriptscriptstyle \pm 0.004}$
        & ${3.286}_{\scriptscriptstyle \pm 0.019}$ \\

        U2Fusion~\cite{Xu2020U2FusionAU}
        & ${3.152}_{\scriptscriptstyle \pm 0.064}$
        & ${0.678}_{\scriptscriptstyle \pm 0.038}$
        & ${0.547}_{\scriptscriptstyle \pm 0.046}$
        & ${\second{0.480}}_{\scriptscriptstyle \pm 0.001}$
        & ${0.513}_{\scriptscriptstyle \pm 0.001}$
        & ${3.518}_{\scriptscriptstyle \pm 0.014}$ \\

        GANMcC~\cite{Ma2021GANMcCAG}
        & ${3.187}_{\scriptscriptstyle \pm 0.086}$
        & ${0.607}_{\scriptscriptstyle \pm 0.018}$
        & ${0.414}_{\scriptscriptstyle \pm 0.072}$
        & ${0.448}_{\scriptscriptstyle \pm 0.011}$
        & ${0.516}_{\scriptscriptstyle \pm 0.009}$
        & ${3.862}_{\scriptscriptstyle \pm 0.114}$ \\

        SDNet~\cite{Zhang2021SDNetAV}
        & ${3.078}_{\scriptscriptstyle \pm 0.086}$
        & ${0.568}_{\scriptscriptstyle \pm 0.048}$
        & ${0.546}_{\scriptscriptstyle \pm 0.009}$
        & ${0.473}_{\scriptscriptstyle \pm 0.014}$
        & ${0.503}_{\scriptscriptstyle \pm 0.011}$
        & ${3.390}_{\scriptscriptstyle \pm 0.004}$ \\

        TarDal~\cite{Liu2022TargetawareDA}
        & ${3.584}_{\scriptscriptstyle \pm 0.015}$
        & ${0.649}_{\scriptscriptstyle \pm 0.003}$
        & ${0.441}_{\scriptscriptstyle \pm 0.006}$
        & ${0.472}_{\scriptscriptstyle \pm 0.002}$
        & ${0.509}_{\scriptscriptstyle \pm 0.001}$
        & ${3.350}_{\scriptscriptstyle \pm 0.022}$ \\

        SegMif~\cite{Liu2023MultiinteractiveFL}
        & ${3.640}_{\scriptscriptstyle \pm 0.380}$
        & ${0.793}_{\scriptscriptstyle \pm 0.016}$
        & ${0.606}_{\scriptscriptstyle \pm 0.061}$
        & ${0.446}_{\scriptscriptstyle \pm 0.005}$
        & ${0.481}_{\scriptscriptstyle \pm 0.032}$
        & ${3.070}_{\scriptscriptstyle \pm 0.016}$ \\

        DDFM~\cite{Zhao2023DDFMDD}
        & ${3.153}_{\scriptscriptstyle \pm 0.005}$
        & ${0.617}_{\scriptscriptstyle \pm 0.069}$
        & ${0.561}_{\scriptscriptstyle \pm 0.000}$
        & ${0.381}_{\scriptscriptstyle \pm 0.035}$
        & ${0.519}_{\scriptscriptstyle \pm 0.002}$
        & ${3.827}_{\scriptscriptstyle \pm 0.006}$ \\

        Dif-Fusion~\cite{Yue2023DifFusionTH}
        & ${3.330}_{\scriptscriptstyle \pm 0.100}$
        & ${0.572}_{\scriptscriptstyle \pm 0.011}$
        & ${0.506}_{\scriptscriptstyle \pm 0.017}$
        & ${0.358}_{\scriptscriptstyle \pm 0.003}$
        & ${0.468}_{\scriptscriptstyle \pm 0.001}$
        & ${3.726}_{\scriptscriptstyle \pm 0.014}$ \\

        DCINN~\cite{Wang2023AGP}
        & ${3.631}_{\scriptscriptstyle \pm 0.014}$
        & ${0.713}_{\scriptscriptstyle \pm 0.039}$
        & ${0.565}_{\scriptscriptstyle \pm 0.020}$
        & ${\second{0.480}}_{\scriptscriptstyle \pm 0.001}$
        & ${\best{0.525}}_{\scriptscriptstyle \pm 0.005}$
        & ${3.526}_{\scriptscriptstyle \pm 0.005}$ \\

        MRFS~\cite{Zhang2024MRFSMR}
        & ${3.288}_{\scriptscriptstyle \pm 0.019}$
        & ${0.604}_{\scriptscriptstyle \pm 0.027}$
        & ${0.508}_{\scriptscriptstyle \pm 0.008}$
        & ${0.334}_{\scriptscriptstyle \pm 0.040}$
        & ${0.419}_{\scriptscriptstyle \pm 0.009}$
        & ${4.092}_{\scriptscriptstyle \pm 0.075}$ \\

        CCF~\cite{cao2024conditional}
        & ${2.923}_{\scriptscriptstyle \pm 0.035}$
        & ${0.507}_{\scriptscriptstyle \pm 0.002}$
        & ${0.397}_{\scriptscriptstyle \pm 0.017}$
        & ${0.366}_{\scriptscriptstyle \pm 0.004}$
        & ${0.500}_{\scriptscriptstyle \pm 0.001}$
        & ${3.280}_{\scriptscriptstyle \pm 0.010}$ \\

        A2RNet~\cite{Li2024A2RNetAA}
        & ${\second{3.666}}_{\scriptscriptstyle \pm 0.075}$
        & ${0.497}_{\scriptscriptstyle \pm 0.009}$
        & ${0.357}_{\scriptscriptstyle \pm 0.007}$
        & ${0.276}_{\scriptscriptstyle \pm 0.032}$
        & ${0.444}_{\scriptscriptstyle \pm 0.023}$
        & ${5.275}_{\scriptscriptstyle \pm 0.016}$ \\

        DiFusionSeg~\cite{Wang2025DiFusionSegDS}
        & ${3.615}_{\scriptscriptstyle \pm 0.119}$
        & ${\best{0.849}}_{\scriptscriptstyle \pm 0.001}$
        & ${\second{0.666}}_{\scriptscriptstyle \pm 0.018}$
        & ${0.431}_{\scriptscriptstyle \pm 0.016}$
        & ${0.462}_{\scriptscriptstyle \pm 0.013}$
        & ${\best{2.598}}_{\scriptscriptstyle \pm 0.043}$ \\

        MaeFuse~\cite{10893688}
        & ${2.404}_{\scriptscriptstyle \pm 0.114}$
        & ${0.605}_{\scriptscriptstyle \pm 0.007}$
        & ${0.486}_{\scriptscriptstyle \pm 0.011}$
        & ${0.399}_{\scriptscriptstyle \pm 0.007}$
        & ${0.500}_{\scriptscriptstyle \pm 0.002}$
        & ${4.356}_{\scriptscriptstyle \pm 0.014}$ \\

        Mask-D.~\cite{11162636}
        & ${3.158}_{\scriptscriptstyle \pm 0.000}$
        & ${0.755}_{\scriptscriptstyle \pm 0.005}$
        & ${0.514}_{\scriptscriptstyle \pm 0.002}$
        & ${0.408}_{\scriptscriptstyle \pm 0.009}$
        & ${0.493}_{\scriptscriptstyle \pm 0.006}$
        & ${3.824}_{\scriptscriptstyle \pm 0.020}$ \\

        \textcolor{red}{OpIVF}~\cite{Xie2024OverexposedIA}
        & ${3.361}_{\scriptscriptstyle \pm 0.027}$
        & ${0.804}_{\scriptscriptstyle \pm 0.003}$
        & ${0.623}_{\scriptscriptstyle \pm 0.002}$
        & ${0.458}_{\scriptscriptstyle \pm 0.002}$
        & ${0.464}_{\scriptscriptstyle \pm 0.001}$
        & ${3.149}_{\scriptscriptstyle \pm 0.027}$ \\

        \textbf{EPOFusion$^{*}$}
        & ${\best{3.711}}_{\scriptscriptstyle \pm 0.021}$
        & ${\second{0.807}}_{\scriptscriptstyle \pm 0.007}$
        & ${\best{0.689}}_{\scriptscriptstyle \pm 0.002}$
        & ${\best{0.484}}_{\scriptscriptstyle \pm 0.000}$
        & ${\second{0.520}}_{\scriptscriptstyle \pm 0.006}$
        & ${\second{2.844}}_{\scriptscriptstyle \pm 0.016}$ \\

        \bottomrule
    \end{tabular*}
    \vspace{-0.3cm}
\end{table}
\begin{table}[t!]
    \centering
    \caption{\textcolor{red}{Quantitative comparisons on 447 image pairs from the IVOE dataset.}}
    \label{IVOE}
    \fontsize{10pt}{12pt}\selectfont
    \setlength{\tabcolsep}{3pt}
    \renewcommand{\arraystretch}{1.1}
    \begin{tabular*}{\columnwidth}{@{\extracolsep{\fill}}l*{6}{>{\centering\arraybackslash}c}}
        \toprule
        \textbf{Category} 
        & MI$\uparrow$ 
        & VIF$\uparrow$ 
        & \(Q^{AB/F}\)$\uparrow$ 
        & SSIM$\uparrow$ 
        & MS\_SSIM$\uparrow$ 
        & PI$\downarrow$ \\
        \midrule

        FusionGAN~\cite{Ma2019FusionGANAG}
        & ${2.589}_{\scriptscriptstyle \pm 0.060}$
        & ${0.419}_{\scriptscriptstyle \pm 0.005}$
        & ${0.278}_{\scriptscriptstyle \pm 0.008}$
        & ${0.352}_{\scriptscriptstyle \pm 0.010}$
        & ${0.465}_{\scriptscriptstyle \pm 0.002}$
        & ${3.966}_{\scriptscriptstyle \pm 0.084}$ \\

        U2Fusion~\cite{Xu2020U2FusionAU}
        & ${2.852}_{\scriptscriptstyle \pm 0.007}$
        & ${0.634}_{\scriptscriptstyle \pm 0.001}$
        & ${0.453}_{\scriptscriptstyle \pm 0.002}$
        & ${0.477}_{\scriptscriptstyle \pm 0.002}$
        & ${0.508}_{\scriptscriptstyle \pm 0.000}$
        & ${3.434}_{\scriptscriptstyle \pm 0.018}$ \\

        GANMcC~\cite{Ma2021GANMcCAG}
        & ${2.807}_{\scriptscriptstyle \pm 0.093}$
        & ${0.523}_{\scriptscriptstyle \pm 0.060}$
        & ${0.257}_{\scriptscriptstyle \pm 0.076}$
        & ${0.412}_{\scriptscriptstyle \pm 0.057}$
        & ${0.508}_{\scriptscriptstyle \pm 0.024}$
        & ${4.344}_{\scriptscriptstyle \pm 0.043}$ \\

        SDNet~\cite{Zhang2021SDNetAV}
        & ${2.505}_{\scriptscriptstyle \pm 0.004}$
        & ${0.513}_{\scriptscriptstyle \pm 0.001}$
        & ${0.482}_{\scriptscriptstyle \pm 0.000}$
        & ${0.471}_{\scriptscriptstyle \pm 0.002}$
        & ${0.512}_{\scriptscriptstyle \pm 0.001}$
        & ${3.028}_{\scriptscriptstyle \pm 0.002}$ \\

        TarDal~\cite{Liu2022TargetawareDA}
        & ${3.020}_{\scriptscriptstyle \pm 0.011}$
        & ${0.622}_{\scriptscriptstyle \pm 0.005}$
        & ${0.456}_{\scriptscriptstyle \pm 0.004}$
        & ${0.447}_{\scriptscriptstyle \pm 0.012}$
        & ${0.513}_{\scriptscriptstyle \pm 0.003}$
        & ${3.455}_{\scriptscriptstyle \pm 0.022}$ \\

        SegMif~\cite{Liu2023MultiinteractiveFL}
        & ${2.685}_{\scriptscriptstyle \pm 0.066}$
        & ${0.684}_{\scriptscriptstyle \pm 0.035}$
        & ${0.502}_{\scriptscriptstyle \pm 0.021}$
        & ${0.452}_{\scriptscriptstyle \pm 0.010}$
        & ${0.506}_{\scriptscriptstyle \pm 0.015}$
        & ${2.775}_{\scriptscriptstyle \pm 0.095}$ \\

        DDFM~\cite{Zhao2023DDFMDD}
        & ${2.828}_{\scriptscriptstyle \pm 0.006}$
        & ${0.599}_{\scriptscriptstyle \pm 0.023}$
        & ${0.389}_{\scriptscriptstyle \pm 0.006}$
        & ${0.431}_{\scriptscriptstyle \pm 0.002}$
        & ${0.503}_{\scriptscriptstyle \pm 0.001}$
        & ${4.110}_{\scriptscriptstyle \pm 0.002}$ \\

        Dif-Fusion~\cite{Yue2023DifFusionTH}
        & ${3.155}_{\scriptscriptstyle \pm 0.041}$
        & ${0.576}_{\scriptscriptstyle \pm 0.005}$
        & ${0.466}_{\scriptscriptstyle \pm 0.003}$
        & ${0.399}_{\scriptscriptstyle \pm 0.001}$
        & ${0.508}_{\scriptscriptstyle \pm 0.001}$
        & ${3.062}_{\scriptscriptstyle \pm 0.010}$ \\

        DCINN~\cite{Wang2023AGP}
        & ${\second{3.250}}_{\scriptscriptstyle \pm 0.022}$
        & ${0.668}_{\scriptscriptstyle \pm 0.023}$
        & ${0.481}_{\scriptscriptstyle \pm 0.014}$
        & ${0.476}_{\scriptscriptstyle \pm 0.001}$
        & ${0.500}_{\scriptscriptstyle \pm 0.004}$
        & ${3.666}_{\scriptscriptstyle \pm 0.032}$ \\

        MRFS~\cite{Zhang2024MRFSMR}
        & ${2.855}_{\scriptscriptstyle \pm 0.006}$
        & ${0.505}_{\scriptscriptstyle \pm 0.024}$
        & ${0.316}_{\scriptscriptstyle \pm 0.026}$
        & ${0.347}_{\scriptscriptstyle \pm 0.022}$
        & ${0.451}_{\scriptscriptstyle \pm 0.018}$
        & ${4.722}_{\scriptscriptstyle \pm 0.034}$ \\

        CCF~\cite{cao2024conditional}
        & ${2.596}_{\scriptscriptstyle \pm 0.001}$
        & ${0.534}_{\scriptscriptstyle \pm 0.000}$
        & ${0.412}_{\scriptscriptstyle \pm 0.002}$
        & ${0.473}_{\scriptscriptstyle \pm 0.003}$
        & ${0.503}_{\scriptscriptstyle \pm 0.003}$
        & ${3.155}_{\scriptscriptstyle \pm 0.042}$ \\

        A2RNet~\cite{Li2024A2RNetAA}
        & ${3.046}_{\scriptscriptstyle \pm 0.009}$
        & ${0.475}_{\scriptscriptstyle \pm 0.020}$
        & ${0.286}_{\scriptscriptstyle \pm 0.026}$
        & ${0.293}_{\scriptscriptstyle \pm 0.033}$
        & ${0.463}_{\scriptscriptstyle \pm 0.033}$
        & ${5.995}_{\scriptscriptstyle \pm 0.078}$ \\

        DiFusionSeg~\cite{Wang2025DiFusionSegDS}
        & ${3.183}_{\scriptscriptstyle \pm 0.038}$
        & ${0.732}_{\scriptscriptstyle \pm 0.037}$
        & ${0.592}_{\scriptscriptstyle \pm 0.010}$
        & ${0.480}_{\scriptscriptstyle \pm 0.008}$
        & ${\best{0.532}}_{\scriptscriptstyle \pm 0.002}$
        & ${2.837}_{\scriptscriptstyle \pm 0.022}$ \\

        MaeFuse~\cite{10893688}
        & ${2.426}_{\scriptscriptstyle \pm 0.038}$
        & ${0.579}_{\scriptscriptstyle \pm 0.003}$
        & ${0.455}_{\scriptscriptstyle \pm 0.003}$
        & ${0.450}_{\scriptscriptstyle \pm 0.000}$
        & ${0.512}_{\scriptscriptstyle \pm 0.001}$
        & ${3.781}_{\scriptscriptstyle \pm 0.082}$ \\

        Mask-D.~\cite{11162636}
        & ${2.602}_{\scriptscriptstyle \pm 0.007}$
        & ${0.657}_{\scriptscriptstyle \pm 0.004}$
        & ${0.463}_{\scriptscriptstyle \pm 0.001}$
        & ${0.458}_{\scriptscriptstyle \pm 0.004}$
        & ${0.510}_{\scriptscriptstyle \pm 0.007}$
        & ${3.049}_{\scriptscriptstyle \pm 0.039}$ \\

        \textcolor{red}{OpIVF}~\cite{Xie2024OverexposedIA}
        & ${3.132}_{\scriptscriptstyle \pm 0.039}$
        & ${0.770}_{\scriptscriptstyle \pm 0.005}$
        & ${0.543}_{\scriptscriptstyle \pm 0.003}$
        & ${0.464}_{\scriptscriptstyle \pm 0.004}$
        & ${0.473}_{\scriptscriptstyle \pm 0.001}$
        & ${2.854}_{\scriptscriptstyle \pm 0.038}$ \\

        \textbf{EPOFusion$^{*}$}
        & ${\best{3.379}}_{\scriptscriptstyle \pm 0.026}$
        & ${\second{0.773}}_{\scriptscriptstyle \pm 0.014}$
        & ${\second{0.598}}_{\scriptscriptstyle \pm 0.010}$
        & ${\best{0.484}}_{\scriptscriptstyle \pm 0.004}$
        & ${\second{0.524}}_{\scriptscriptstyle \pm 0.002}$
        & ${\best{2.468}}_{\scriptscriptstyle \pm 0.025}$ \\

        \textbf{EPOFusion}
        & ${3.171}_{\scriptscriptstyle \pm 0.008}$
        & ${\best{0.782}}_{\scriptscriptstyle \pm 0.005}$
        & ${\best{0.601}}_{\scriptscriptstyle \pm 0.007}$
        & ${\second{0.482}}_{\scriptscriptstyle \pm 0.003}$
        & ${0.513}_{\scriptscriptstyle \pm 0.002}$
        & ${\second{2.751}}_{\scriptscriptstyle \pm 0.009}$ \\

        \bottomrule
    \end{tabular*}
    \vspace{-0.3cm}
\end{table}
\begin{table}[htbp]
    \vspace{-0.3cm}
    \centering
    \caption{\textcolor{red}{
    Paired statistical analysis of per-image results averaged across seeds 0, 1, and 2 on MSRS, FMB, and IVOE. Rank-1 and Rank-2 denote the first- and second-ranked methods for each metric and dataset.
    }}
    \label{statistical_analysis}
    \fontsize{10pt}{12pt}\selectfont
    \renewcommand{\arraystretch}{1.2}
    \setlength{\tabcolsep}{6pt}
    \begin{tabular}{lllcc}
        \toprule
        Dataset & Metric & Comparison & 95\% CI & $p$-value \\
        \midrule

        MSRS & VIF
        & EPOFusion$^{*}$ vs. Rank-2
        & $[-0.001,\,+0.028]$ 
        & $0.063$ \\

        MSRS & $Q^{AB/F}$
        & EPOFusion$^{*}$ vs. Rank-2
        & $[+0.026,\,+0.032]$
        & $<0.001$ \\

        MSRS & SSIM
        & EPOFusion$^{*}$ vs. Rank-2
        & $[+0.002,\,+0.006]$
        & $<0.001$ \\
        \midrule

        FMB & VIF
        & EPOFusion$^{*}$ vs. Rank-1
        & $[-0.047,\,-0.037]$
        & $<0.001$ \\

        FMB & $Q^{AB/F}$
        & EPOFusion$^{*}$ vs. Rank-2
        & $[+0.021,\,+0.023]$
        & $<0.001$ \\

        FMB & SSIM
        & EPOFusion$^{*}$ vs. Rank-2
        & $[+0.002,\,+0.005]$
        & $<0.001$ \\
        \midrule

        IVOE & VIF
        & EPOFusion vs. EPOFusion$^{*}$
        & $[+0.002,\,+0.018]$
        & $<0.05$ \\

        IVOE & $Q^{AB/F}$
        & EPOFusion vs. EPOFusion$^{*}$
        & $[+0.002,\,+0.008]$
        & $<0.05$ \\

        IVOE & SSIM
        & EPOFusion vs. EPOFusion$^{*}$
        & $[-0.002,\,+0.001]$
        & $0.299$ \\

        \bottomrule
    \end{tabular}
    \vspace{-0.3cm}
\end{table}
\begin{figure*}[t]
    \centering
    \includegraphics[width=1.00\textwidth]{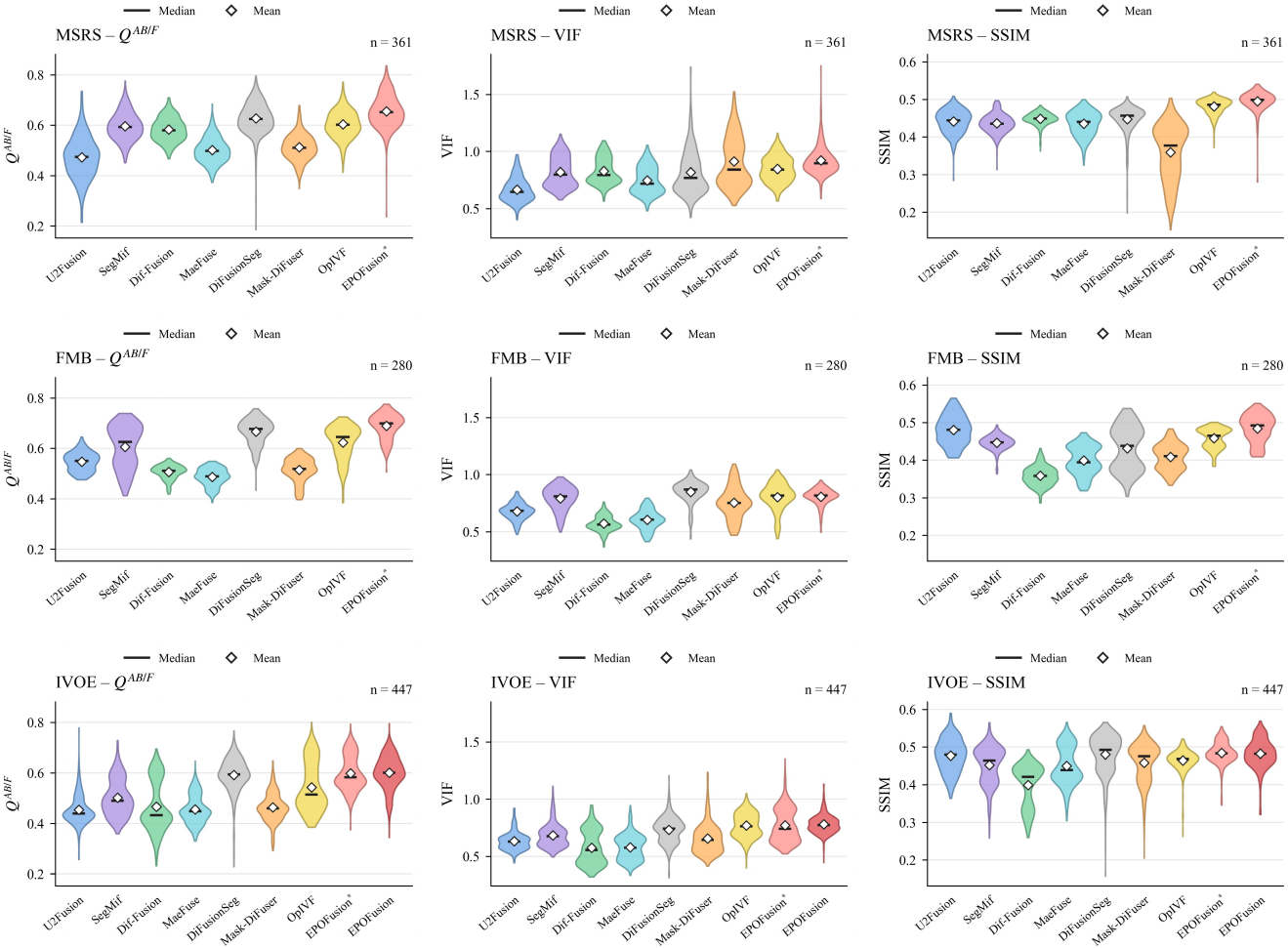}
    \caption{\textcolor{red}{Violin plots of per-image $Q^{AB/F}$, VIF, and SSIM for representative fusion methods on MSRS, FMB, and IVOE, averaged over seeds 0, 1, and 2.}}
    \label{Violin}
    \vspace{-0.6cm}
\end{figure*}
\section{Experiments}\label{sec4}
This section introduces the experimental settings and metrics, compares EPOFusion with existing methods, analyzes its computational complexity, and evaluates its downstream task performance.
\subsection{Experimental Setup}
\subsubsection{Implementation Details}
All experiments, including comparative and ablation studies, are performed on a system running Ubuntu 20.04 with the PyTorch 2.2.0 framework. The hardware setup includes an Intel(R) Core(TM) i7-13700KF processor operating at 3.4 GHz, 64GB of RAM, and an NVIDIA GeForce RTX 4090 GPU with 24GB of VRAM. For optimization, the AdamW optimizer~\cite{Loshchilov2017DecoupledWD} is used, with a linear learning rate schedule and a warm-up phase. The maximum learning rate is set to \(1 \times 10^{-5}\), the batch size is 8. \textcolor{red}{All input images are resized to $480\times320$ and normalized to $[0,1]$. Unless otherwise stated, EPOFusion uses three refinement steps during inference. The final-epoch checkpoint is used for evaluation.}
\subsubsection{Evaluation Metrics}
To comprehensively evaluate fusion performance, we employ a diverse set of metrics, including objective measures like mutual information (MI) calculated in the fused image, source referenced metrics such as $Q^{AB/F}$, structural similarity (SSIM), multi-scale structural similarity (MS-SSIM), the visual information fidelity (VIF), and perceptual index (PI)~\cite{blau20182018}. For all metrics except PI, higher values indicate better performance, while lower PI values indicate better perceptual quality.

\begin{figure*}[!t]
  \centering
  \subfloat[\scriptsize Day-time street scene]{\includegraphics[width=0.50\textwidth]{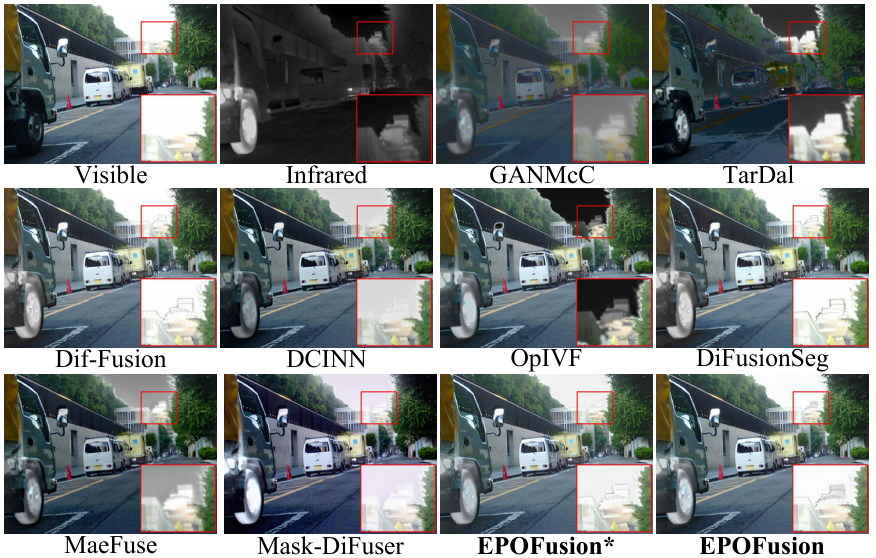}\label{DayScene}}
  \subfloat[\scriptsize Night-time street scene]{\includegraphics[width=0.50\textwidth]{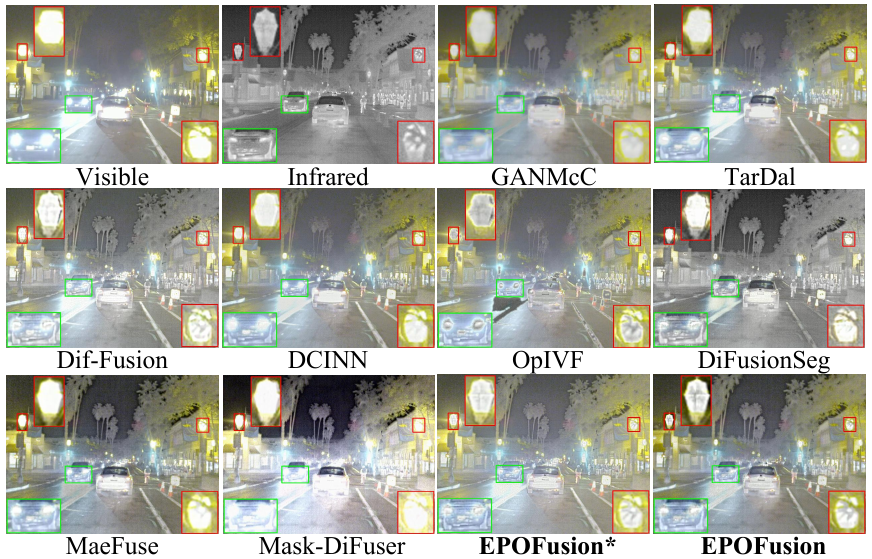}\label{NightScene}}
  \\
  \subfloat[\scriptsize Night-time driving scene with visible overexposure]{\includegraphics[width=0.50\textwidth]{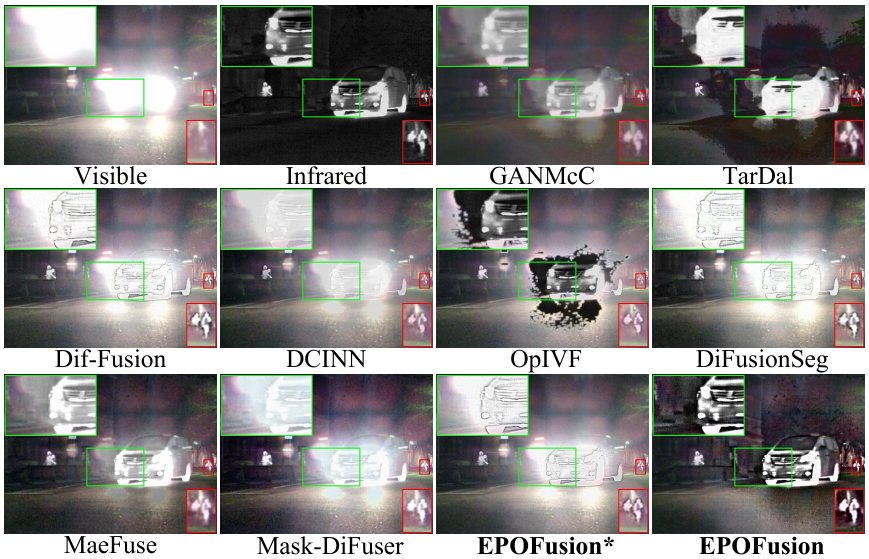}\label{NightVi}}
  \subfloat[\scriptsize Scene with intensified infrared thermal emission]{\includegraphics[width=0.50\textwidth]{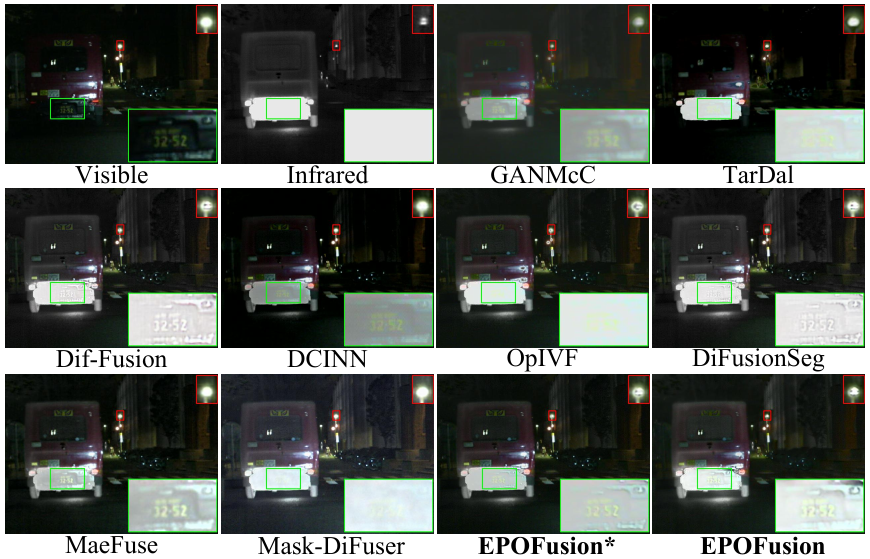}\label{NightIr}}
  \caption{Fusion results in various scenarios, with visible and infrared source images, outputs from several SOTA methods and our method, and enlarged regions highlighting details.}
\label{FusionResult}
\vspace{-0.6cm}
\end{figure*}

\subsection{Fusion Comparison and Analysis}
To evaluate the fusion performance of EPOFusion, we conduct experiments on MSRS~\cite{Tang2022ImageFI}, FMB~\cite{Liu2023MultiinteractiveFL}, and the proposed IVOE dataset. \textcolor{red}{Except for the training-free methods CCF~\cite{cao2024conditional} and DDFM~\cite{Zhao2023DDFMDD}, all competing methods are initialized from their officially released pretrained weights and fine-tuned for 30 epochs on the corresponding training splits following their official training configurations. For each dataset, EPOFusion$^{*}$ is trained on the same corresponding training split and evaluated on the same test split as the competing methods. Each experiment is independently repeated with three random seeds (0, 1, and 2), and the mean and standard deviation are reported.} To ensure a fair comparison of fusion performance, we report EPOFusion$^{*}$, \textcolor{red}{which removes the exposure-aware guidance and region-aware fusion supervision and clean-reference supervision while retaining the same progressive fusion architecture. The full EPOFusion is reported only on IVOE to demonstrate the additional benefits of the exposure-aware compensation strategy.}

\subsubsection{Quantitative Comparison and Analysis}
We compare EPOFusion with 16 SOTA fusion algorithms from recent years, and the quantitative results are presented in~\autoref{MSRS},~\autoref{FMB}, and~\autoref{IVOE} on MSRS, FMB, and IVOE datasets, respectively. On the two public datasets, EPOFusion$^{*}$ demonstrates strong fusion performance. \textcolor{red}{Specifically, on MSRS, EPOFusion$^{*}$ achieves the best results on MI, VIF, $Q^{AB/F}$, and SSIM, with scores of 3.615, 0.921, 0.655, and 0.495, respectively, while ranking second on MS\_SSIM with a score of 0.521. On FMB, EPOFusion$^{*}$ achieves the best results on MI, $Q^{AB/F}$, and SSIM, with scores of 3.711, 0.689, and 0.484, respectively, and ranks second on VIF, MS\_SSIM, and PI with scores of 0.807, 0.520, and 2.844.} MI measures the amount of information transferred from the source images to the fused result, VIF evaluates visual information fidelity, and $Q^{AB/F}$ quantifies the preservation of edge and gradient information. \textcolor{red}{These results demonstrate that EPOFusion$^{*}$ effectively preserves multimodal information, structural details, and perceptual quality.}

On the IVOE dataset, EPOFusion$^{*}$ remains highly competitive, \textcolor{red}{achieving the best MI, SSIM, and PI scores of 3.379, 0.484, and 2.468, respectively, while ranking second on VIF, $Q^{AB/F}$, and MS\_SSIM. With the introduction of exposure-aware guidance, the full EPOFusion further improves VIF from 0.773 to 0.782 and $Q^{AB/F}$ from 0.598 to 0.601, achieving the best performance on both metrics. Meanwhile, it maintains the second-best SSIM and PI scores of 0.482 and 2.751, respectively.}

\textcolor{red}{EPOFusion incorporates an exposure-aware mechanism and a region-aware loss to actively suppress corrupted visible information in overexposed regions and preserve more reliable infrared structural information, thereby preserving more informative content in regions affected by signal saturation. Meanwhile, the enhanced integration of infrared information improves the preservation of effective visual information and edge structures, leading to further gains in VIF and $Q^{AB/F}$. The practical advantages brought by exposure-aware fusion are further evaluated in the subsequent qualitative comparisons and downstream perception evaluations.}

\textcolor{red}{To examine the per-image performance distributions,~\autoref{Violin} presents the distributions of $Q^{AB/F}$, VIF, and SSIM for representative methods on the three datasets. EPOFusion$^{*}$ generally occupies the higher ranges on MSRS and FMB, with particularly high mean and median values for $Q^{AB/F}$ and SSIM. On IVOE, the full EPOFusion further shifts the distribution centers of VIF and $Q^{AB/F}$ upward. These show that the advantages of EPOFusion are reflected in the image-level performance distributions.}

\textcolor{red}{~\autoref{statistical_analysis} further presents paired statistical analysis based on per-image results. On MSRS, the improvements of EPOFusion$^{*}$ over the second-best method in $Q^{AB/F}$ and SSIM are statistically significant, with the corresponding 95\% confidence intervals entirely above zero; the same trend is observed for $Q^{AB/F}$ and SSIM on FMB. On IVOE, the full EPOFusion achieves significant improvements over EPOFusion$^{*}$ in VIF and $Q^{AB/F}$, further validating the contribution of the exposure-aware compensation mechanism to preserving effective visual information and structures.}

\subsubsection{Qualitative Comparison and Analysis}
To better evaluate performance in overexposed scenes, we compare representative fusion approaches in~\autoref{FusionResult}. In~\autoref{DayScene}, generative prior-based methods such as GANMcC~\cite{Ma2021GANMcCAG} may introduce redundant infrared responses because they cannot reliably distinguish useful infrared information, while loss- or task-guided methods such as DCINN~\cite{Wang2023AGP} and DiFusionSeg~\cite{Wang2025DiFusionSegDS} may lose information or retain only partial texture or intensity cues. \textcolor{red}{OpIVF enhances infrared information across the bright region, causing an infrared-dominant appearance in some areas.} \textcolor{red}{EPOFusion$^{*}$} progressively integrates useful infrared and visible information. \textcolor{red}{With exposure-aware compensation}, EPOFusion better preserves infrared structures and intensity while maintaining visual consistency throughout the image.

~\autoref{NightScene} and~\autoref{NightVi} are nighttime scenes with local overexposure caused by vehicle headlights, \textcolor{red}{while infrared provides important complementary cues. In~\autoref{NightScene}, most methods retain partial vehicle textures, but strong illumination weakens local details. In the red-box region, they often fail to recover both the lamp contour and surrounding foliage. EPOFusion$^{*}$ better preserves vehicle textures, the lamp contour, and part of the foliage. EPOFusion further balances exposure and infrared information, preserving clearer lamp details and more complete foliage structures}

\textcolor{red}{In~\autoref{NightVi}, overexposure from oncoming headlights causes substantial information loss in both the vehicle and weak-textured regions.} Although some methods emphasize infrared vehicle textures, strong illumination still prevents sufficient recovery of buildings, vegetation, and pedestrians. \textcolor{red}{\textcolor{red}{GANMcC~\cite{Ma2021GANMcCAG} and MaeFuse~\cite{10893688} partially suppress the excessive brightness, yet the obscured background information remains largely unrecovered.} EPOFusion$^{*}$ preserves the vehicle details and balances overexposure suppression and infrared preservation, preserving more background information while retaining the vehicle structure. In the red-box region, the pedestrian is more visible, with clearer building contours and vegetation textures.}

\autoref{NightIr} presents a challenging degradation scenario with visible underexposure and local infrared saturation. In this case, most competing methods suffer from information loss or blurred details in bright infrared regions, making it difficult to effectively integrate visible details with infrared structural information. \textcolor{red}{Since the overexposure prior of OpIVF~\cite{Xie2024OverexposedIA} mainly targets bright degraded regions in the visible modality, it provides limited guidance when the degradation is dominated by infrared saturation.} In contrast, both EPOFusion$^{*}$ and EPOFusion preserve license-plate information and surrounding details in high-intensity infrared regions, making the plate characters clearer and more distinguishable. In addition, distant light-source structures are better preserved, resulting in richer details and clearer overall structures.

Across four representative overexposure scenarios, \textcolor{red}{EPOFusion effectively alleviates information loss and highlight interference caused by local overexposure while preserving informative infrared structures}, achieving a \textcolor{red}{good balance} between visual quality and information completeness.

\begin{figure*}[t]
    \centering
    \includegraphics[width=1.00\textwidth]{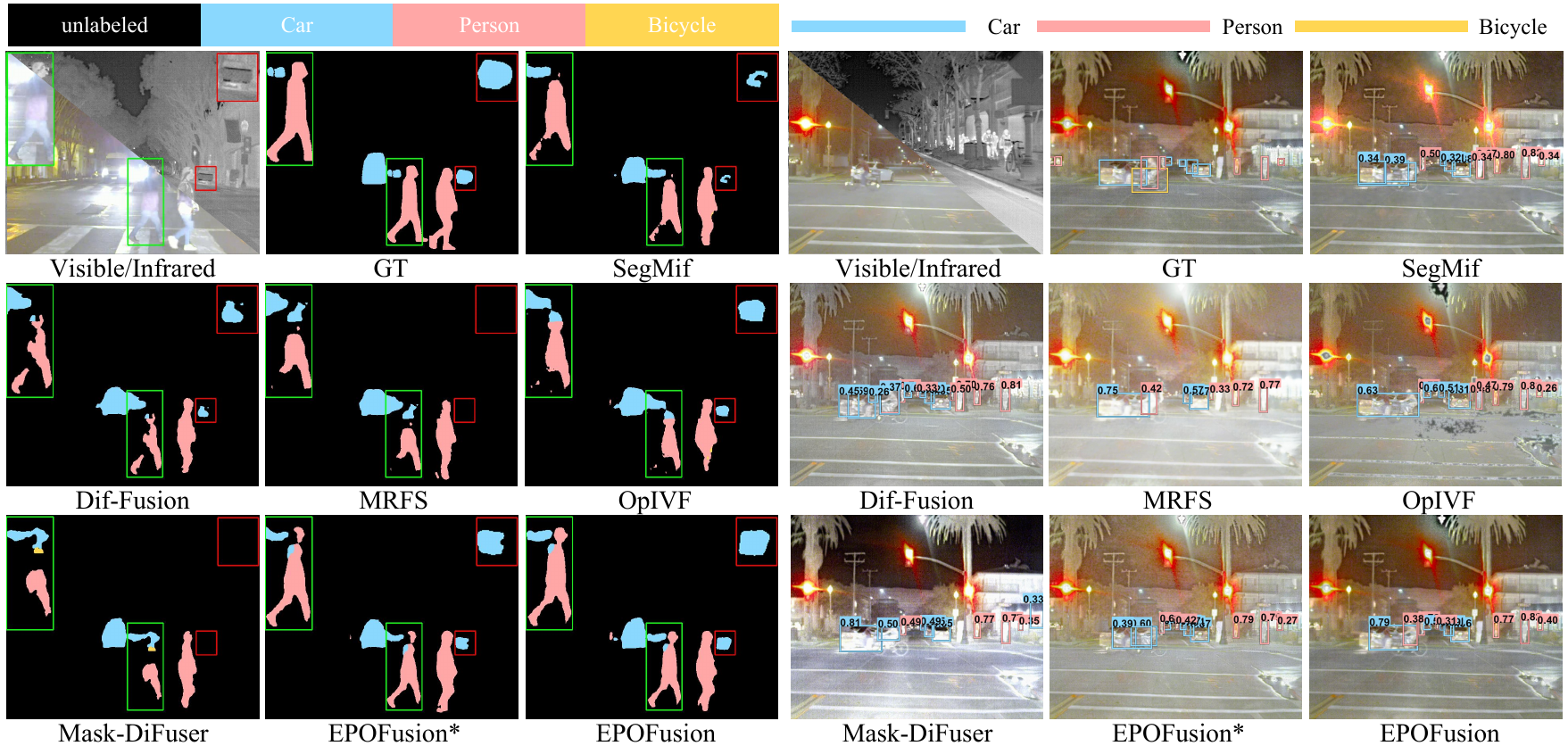}
    \caption{\textcolor{red}{Qualitative comparison of segmentation and detection results on the test split of the real-world IVOE subset}}
    \label{Seg_Detect}
    \vspace{-0.3cm}
\end{figure*}

\subsection{Performance on downstream tasks}
To evaluate the effectiveness of our method in handling overexposed scenarios for downstream tasks, we conduct \textcolor{red}{segmentation and detection experiments on the real-world subset of the IVOE dataset}. \textcolor{red}{Downstream perception tasks offer a more direct assessment of the fusion capability of different methods under overexposed conditions.} SegFormer-B1~\cite{Xie2021SegFormerSA} and YOLOv11m~\cite{Khanam2024YOLOv11AO} are adopted for segmentation and detection, respectively. \textcolor{red}{For downstream evaluation, these real-world samples are further divided into 357 training pairs and 90 test pairs for training and evaluating the downstream perception models, while the fusion network remains fixed. Each fusion method generates fused images for both splits, and the downstream models are trained identically with seeds 0, 1, and 2.}

\subsubsection{Segmentation Comparison and Analysis}
\textcolor{red}{As shown in~\autoref{SegmentationTable}, EPOFusion achieves the best overall segmentation performance, with an mDice of 73.26\% and an mIoU of 60.78\%. At the category level, it obtains the highest Bicycle IoU of 32.72\% and the second-highest Car IoU of 84.42\%, while remaining competitive on People. DiFusionSeg performs well on People through joint fusion--segmentation optimization, U2Fusion achieves strong performance on Car, and GANMcC preserves useful infrared target information. EPOFusion$^{*}$ ranks second overall, achieving an mDice of 72.65\% and an mIoU of 60.23\%, indicating that progressive feature refinement alone can effectively retain useful information in overexposed regions. The full EPOFusion further improves both metrics, demonstrating the additional benefit of the exposure-aware compensation strategy.}

\textcolor{red}{We further provide qualitative comparisons in~\autoref{Seg_Detect}. EPOFusion preserves more complete information in overexposed regions. EPOFusion identifies regions requiring compensation and progressively integrates informative infrared cues through iterative refinement, thereby reducing missed regions and improving target completeness.}

\subsubsection{Detection Comparison and Analysis}
\textcolor{red}{As shown in~\autoref{DetectionTable}, EPOFusion achieves the best overall detection performance, with mAP${50}$ and mAP${50:95}$ scores of 63.03\% and 31.52\%, respectively. At the category level, it achieves the highest AP${50}$ scores on People and Bicycle among the compared fusion methods, reaching 71.24\% and 37.91\%, respectively, while maintaining competitive performance on Car with an AP${50}$ of 79.93\%. EPOFusion$^{*}$ achieves mAP${50}$ and mAP${50:95}$ scores of 61.02\% and 30.40\%, respectively. As shown in~\autoref{Seg_Detect}, both EPOFusion and EPOFusion$^{*}$ preserve more complete target information and reduce missed detections in strongly saturated and locally overexposed regions, while the full EPOFusion shows more complete target preservation in degraded areas.}

These results indicate that preserving informative infrared structures benefits both visual quality and downstream perception under real-world overexposure.

\begin{table}[t!]
    \centering
    \caption{\textcolor{red}{Comparison of segmentation performance on the test split of the real-world subset of IVOE.}}
    \label{SegmentationTable}
    \fontsize{10pt}{12pt}\selectfont
    \setlength{\tabcolsep}{3pt}
    \renewcommand{\arraystretch}{1.1}
    \begin{tabular*}{\columnwidth}{@{\extracolsep{\fill}}l*{5}{>{\centering\arraybackslash}c}}
        \toprule
        \multirow{2}{*}{\textbf{Method}}
        & \multicolumn{3}{c}{IoU$\uparrow$}
        & \multirow{2}{*}{mDice$\uparrow$}
        & \multirow{2}{*}{mIoU$\uparrow$} \\
        \cmidrule(lr){2-4}
        & People & Car & Bicycle & & \\
        \midrule

        VI
        & ${43.07}_{\scriptstyle \pm 0.12}$
        & ${80.88}_{\scriptstyle \pm 0.15}$
        & ${27.45}_{\scriptstyle \pm 1.49}$
        & ${64.24}_{\scriptstyle \pm 0.64}$
        & ${50.47}_{\scriptstyle \pm 0.52}$ \\

        IR
        & ${67.19}_{\scriptstyle \pm 1.01}$
        & ${82.54}_{\scriptstyle \pm 1.00}$
        & ${29.14}_{\scriptstyle \pm 3.01}$
        & ${71.99}_{\scriptstyle \pm 1.06}$
        & ${59.62}_{\scriptstyle \pm 0.94}$ \\

        \midrule

        GANMcC\textsubscript{20}
        & ${62.51}_{\scriptstyle \pm 0.62}$
        & ${84.15}_{\scriptstyle \pm 0.49}$
        & ${28.34}_{\scriptstyle \pm 2.33}$
        & ${70.82}_{\scriptstyle \pm 1.04}$
        & ${58.34}_{\scriptstyle \pm 0.92}$ \\

        U2Fusion\textsubscript{20}
        & ${63.35}_{\scriptstyle \pm 1.06}$
        & $\best{84.71}_{\scriptstyle \pm 0.47}$
        & ${27.62}_{\scriptstyle \pm 1.41}$
        & ${70.85}_{\scriptstyle \pm 0.64}$
        & ${58.56}_{\scriptstyle \pm 0.64}$ \\

        TarDal\textsubscript{22}
        & ${58.46}_{\scriptstyle \pm 0.36}$
        & ${82.05}_{\scriptstyle \pm 0.33}$
        & ${21.74}_{\scriptstyle \pm 1.78}$
        & ${66.54}_{\scriptstyle \pm 0.65}$
        & ${54.08}_{\scriptstyle \pm 0.39}$ \\

        SegMif\textsubscript{23}
        & ${60.57}_{\scriptstyle \pm 0.17}$
        & ${83.37}_{\scriptstyle \pm 0.90}$
        & ${26.54}_{\scriptstyle \pm 1.11}$
        & ${69.44}_{\scriptstyle \pm 0.50}$
        & ${56.83}_{\scriptstyle \pm 0.49}$ \\

        Dif-Fusion\textsubscript{23}
        & ${63.88}_{\scriptstyle \pm 0.18}$
        & ${83.47}_{\scriptstyle \pm 0.78}$
        & ${25.44}_{\scriptstyle \pm 1.84}$
        & ${69.83}_{\scriptstyle \pm 0.64}$
        & ${57.60}_{\scriptstyle \pm 0.39}$ \\

        MRFS\textsubscript{24}
        & ${59.04}_{\scriptstyle \pm 0.95}$
        & ${81.25}_{\scriptstyle \pm 1.77}$
        & ${28.44}_{\scriptstyle \pm 0.46}$
        & ${69.39}_{\scriptstyle \pm 0.57}$
        & ${56.25}_{\scriptstyle \pm 0.75}$ \\

        DiFusionSeg\textsubscript{25}
        & $\best{66.11}_{\scriptstyle \pm 0.58}$
        & ${82.37}_{\scriptstyle \pm 2.39}$
        & ${29.95}_{\scriptstyle \pm 6.12}$
        & ${71.92}_{\scriptstyle \pm 2.13}$
        & ${59.48}_{\scriptstyle \pm 1.53}$ \\

        Mask-DiFuser\textsubscript{26}
        & ${60.53}_{\scriptstyle \pm 0.98}$
        & ${82.57}_{\scriptstyle \pm 0.43}$
        & ${30.22}_{\scriptstyle \pm 1.77}$
        & ${70.75}_{\scriptstyle \pm 0.98}$
        & ${57.77}_{\scriptstyle \pm 0.99}$ \\

        OpIVF\textsubscript{24}
        & ${63.49}_{\scriptstyle \pm 0.19}$
        & ${82.17}_{\scriptstyle \pm 0.96}$
        & ${27.07}_{\scriptstyle \pm 3.81}$
        & ${70.14}_{\scriptstyle \pm 1.83}$
        & ${57.58}_{\scriptstyle \pm 1.63}$ \\

        \midrule

        \textbf{EPOFusion$^{*}$}
        & $\second{65.95}_{\scriptstyle \pm 2.06}$
        & ${83.60}_{\scriptstyle \pm 0.13}$
        & $\second{31.13}_{\scriptstyle \pm 3.02}$
        & $\second{72.65}_{\scriptstyle \pm 0.86}$
        & $\second{60.23}_{\scriptstyle \pm 0.67}$ \\

        \textbf{EPOFusion}
        & ${65.21}_{\scriptstyle \pm 1.23}$
        & $\second{84.42}_{\scriptstyle \pm 0.68}$
        & $\best{32.72}_{\scriptstyle \pm 2.28}$
        & $\best{73.26}_{\scriptstyle \pm 0.49}$
        & $\best{60.78}_{\scriptstyle \pm 0.27}$ \\

        \bottomrule
    \end{tabular*}
    \vspace{-0.3cm}
\end{table}

\begin{table}[t!]
    \centering
    \caption{\textcolor{red}{Comparison of detection performance on the test split of the real-world subset of IVOE.}}
    \label{DetectionTable}
    \fontsize{10pt}{12pt}\selectfont
    \setlength{\tabcolsep}{3pt}
    \renewcommand{\arraystretch}{1.1}
    \begin{tabular*}{\columnwidth}{@{\extracolsep{\fill}}l*{5}{>{\centering\arraybackslash}c}}
        \toprule
        \multirow{2}{*}{\textbf{Method}}
        & \multicolumn{3}{c}{AP$_{50}\uparrow$}
        & \multirow{2}{*}{mAP$_{50}\uparrow$}
        & \multirow{2}{*}{mAP$_{50:95}\uparrow$} \\
        \cmidrule(lr){2-4}
        & People & Car & Bicycle & & \\
        \midrule

        VI
        & ${35.86}_{\scriptstyle \pm 2.06}$
        & ${54.52}_{\scriptstyle \pm 1.04}$
        & ${30.39}_{\scriptstyle \pm 2.87}$
        & ${40.26}_{\scriptstyle \pm 0.51}$
        & ${17.84}_{\scriptstyle \pm 0.56}$ \\

        IR
        & ${71.40}_{\scriptstyle \pm 1.65}$
        & ${78.58}_{\scriptstyle \pm 1.62}$
        & ${32.80}_{\scriptstyle \pm 0.73}$
        & ${60.93}_{\scriptstyle \pm 0.16}$
        & ${29.82}_{\scriptstyle \pm 0.57}$ \\

        \midrule

        GANMcC\textsubscript{20}
        & ${70.07}_{\scriptstyle \pm 0.51}$
        & ${78.40}_{\scriptstyle \pm 0.80}$
        & ${35.16}_{\scriptstyle \pm 1.57}$
        & $\second{61.21}_{\scriptstyle \pm 0.59}$
        & ${30.31}_{\scriptstyle \pm 0.35}$ \\

        U2Fusion\textsubscript{20}
        & ${70.38}_{\scriptstyle \pm 0.90}$
        & $\best{79.99}_{\scriptstyle \pm 1.17}$
        & ${30.33}_{\scriptstyle \pm 3.12}$
        & ${60.23}_{\scriptstyle \pm 0.36}$
        & ${30.31}_{\scriptstyle \pm 0.93}$ \\

        TarDal\textsubscript{22}
        & ${69.34}_{\scriptstyle \pm 0.47}$
        & ${78.10}_{\scriptstyle \pm 0.58}$
        & ${32.38}_{\scriptstyle \pm 3.51}$
        & ${59.94}_{\scriptstyle \pm 1.08}$
        & ${30.20}_{\scriptstyle \pm 0.27}$ \\

        SegMif\textsubscript{23}
        & ${65.89}_{\scriptstyle \pm 0.47}$
        & ${79.52}_{\scriptstyle \pm 1.29}$
        & ${34.00}_{\scriptstyle \pm 0.54}$
        & ${59.80}_{\scriptstyle \pm 0.72}$
        & ${29.68}_{\scriptstyle \pm 0.23}$ \\

        Dif-Fusion\textsubscript{23}
        & $\second{70.41}_{\scriptstyle \pm 0.96}$
        & ${77.71}_{\scriptstyle \pm 0.07}$
        & ${34.25}_{\scriptstyle \pm 2.51}$
        & ${60.79}_{\scriptstyle \pm 0.55}$
        & ${30.24}_{\scriptstyle \pm 1.00}$ \\

        MRFS\textsubscript{24}
        & ${68.05}_{\scriptstyle \pm 1.04}$
        & ${75.03}_{\scriptstyle \pm 1.06}$
        & ${32.43}_{\scriptstyle \pm 2.13}$
        & ${58.50}_{\scriptstyle \pm 1.15}$
        & ${29.17}_{\scriptstyle \pm 0.30}$ \\

        DiFusionSeg\textsubscript{25}
        & ${70.34}_{\scriptstyle \pm 0.24}$
        & ${76.75}_{\scriptstyle \pm 0.95}$
        & ${35.15}_{\scriptstyle \pm 2.00}$
        & ${60.75}_{\scriptstyle \pm 0.46}$
        & ${30.36}_{\scriptstyle \pm 0.11}$ \\

        Mask-DiFuser\textsubscript{26}
        & ${70.27}_{\scriptstyle \pm 0.56}$
        & ${76.20}_{\scriptstyle \pm 0.97}$
        & ${32.16}_{\scriptstyle \pm 4.28}$
        & ${59.54}_{\scriptstyle \pm 1.15}$
        & ${29.87}_{\scriptstyle \pm 0.14}$ \\

        OpIVF\textsubscript{24}
        & ${70.09}_{\scriptstyle \pm 0.69}$
        & ${75.90}_{\scriptstyle \pm 0.64}$
        & ${31.51}_{\scriptstyle \pm 4.94}$
        & ${59.17}_{\scriptstyle \pm 1.31}$
        & ${29.57}_{\scriptstyle \pm 0.10}$ \\

        \midrule

        \textbf{EPOFusion$^{*}$}
        & ${69.66}_{\scriptstyle \pm 0.85}$
        & ${78.16}_{\scriptstyle \pm 0.28}$
        & $\second{35.25}_{\scriptstyle \pm 2.14}$
        & ${61.02}_{\scriptstyle \pm 0.42}$
        & $\second{30.40}_{\scriptstyle \pm 0.44}$ \\

        \textbf{EPOFusion}
        & $\best{71.24}_{\scriptstyle \pm 0.23}$
        & $\second{79.93}_{\scriptstyle \pm 1.22}$
        & $\best{37.91}_{\scriptstyle \pm 2.63}$
        & $\best{63.03}_{\scriptstyle \pm 0.76}$
        & $\best{31.52}_{\scriptstyle \pm 0.42}$ \\

        \bottomrule
    \end{tabular*}
    \vspace{-0.3cm}
\end{table}

\begin{table}[htbp]
    \centering
    \caption{\textcolor{black}{Comparison of different methods in terms of FLOPs, Params, and Time on the MSRS dataset.}}
    \label{complexity}
    \fontsize{10pt}{12pt}\selectfont
    \renewcommand{\arraystretch}{1.2}
    \setlength{\tabcolsep}{4pt}
    \begin{tabular}{lcccc}
        \toprule
        Method & Input Size & Params/M$\downarrow$ & FLOPs/G$\downarrow$ & Time/ms$\downarrow$ \\
        \midrule
        U2Fusion\textsubscript{20}\cite{Xu2020U2FusionAU}     & 480$\times$640 & 0.66   & 518     & 695.64 \\
        SDNet\textsubscript{21}\cite{Zhang2021SDNetAV}        & 480$\times$640 & 0.07   & 64.55   & 122.33 \\
        DDFM\textsubscript{23}\cite{Zhao2023DDFMDD}           & 480$\times$640 & 552.81 & 522052  & 102760 \\
        Dif-Fusion\textsubscript{23}\cite{Yue2023DifFusionTH} & 480$\times$640 & 416.47 & 1056.86 & 809.41 \\
        SegMiF\textsubscript{23}\cite{Liu2023MultiinteractiveFL} & 480$\times$640 & 45.63 & 359.41 & 309.70 \\
        MRFS\textsubscript{24}\cite{Zhang2024MRFSMR}          & 480$\times$640 & 134.96 & 139.13  & 124.72 \\
        A2RNet\textsubscript{25}\cite{Li2024A2RNetAA}         & 480$\times$640 & 3.57   & 164.82  & 239.58 \\
        MaeFuse\textsubscript{25}\cite{10893688}              & 480$\times$640 & 325.02 & 1004.55 & 166.20 \\
        Mask-DiFuser\textsubscript{26}\cite{11162636}         & 480$\times$640 & 168.49 & 4552.29 & 11628 \\
        \textbf{EPOFusion}                                    & 480$\times$640 & 33.87  & 149.99  & 74.80 \\
        \bottomrule
    \end{tabular}
    \vspace{-0.3cm}
\end{table}
\subsection{Complexity Discussion}
We compare EPOFusion with state-of-the-art methods in terms of FLOPs, parameters, and inference time. As summarized in~\autoref{complexity}, our model maintains a moderate number of parameters (33.87M) and achieves fast inference (74.80 ms), striking a favorable balance between fusion quality and efficiency.
This efficiency mainly benefits from the adoption of a DDIM-style iterative update scheme and progressive refinement in the feature space rather than the pixel space, which significantly reduces computational overhead. 
Overall, the proposed method demonstrates a good trade-off between performance and efficiency, showing its potential for practical perception systems.

\begin{table}[t!]
    \centering
    \caption{\textcolor{red}{Ablation studies of EPOFusion on the IVOE dataset. IFR, HFB, and TDE denote iterative feature refinement, high-fidelity branch, and time-dependent encoding, respectively. 'Clean Ref.'' and 'Rule Prior'' denote the clean feature-state reference and threshold-based compensation prior, respectively.}}
    \label{AblationTable}
    \fontsize{10pt}{12pt}\selectfont
    \renewcommand{\arraystretch}{1.2}
    \begin{tabular}{>{\centering\arraybackslash}p{2.5cm}*{5}{>{\centering\arraybackslash}p{1.30cm}}}
        \toprule
        \multicolumn{6}{c}{\textbf{Ablation of Progressive Fusion Framework}} \\
        \midrule
        \textbf{Category} 
        & MI$\uparrow$ 
        & VIF$\uparrow$ 
        & $Q^{AB/F}\uparrow$ 
        & SSIM$\uparrow$
        & mIoU$\uparrow$ \\
        \midrule
        
        w/o IFR 
        & ${\second{2.992}}_{\scriptstyle \pm 0.027}$ 
        & ${0.641}_{\scriptstyle \pm 0.003}$ 
        & ${0.548}_{\scriptstyle \pm 0.006}$ 
        & ${\best{0.487}}_{\scriptstyle \pm 0.001}$ 
        & ${\second{58.07}}_{\scriptstyle \pm 0.19}$ \\
        
        w/o HFB
        & ${2.798}_{\scriptstyle \pm 0.005}$ 
        & ${0.533}_{\scriptstyle \pm 0.003}$ 
        & ${0.468}_{\scriptstyle \pm 0.001}$ 
        & ${0.435}_{\scriptstyle \pm 0.004}$ 
        & ${57.92}_{\scriptstyle \pm 0.31}$ \\
        
        w/o TDE
        & ${2.984}_{\scriptstyle \pm 0.018}$ 
        & ${\second{0.659}}_{\scriptstyle \pm 0.009}$ 
        & ${\second{0.554}}_{\scriptstyle \pm 0.024}$ 
        & ${0.480}_{\scriptstyle \pm 0.002}$ 
        & ${56.76}_{\scriptstyle \pm 0.72}$ \\
        
        EPOFusion$^{*}$
        & ${\best{3.379}}_{\scriptstyle \pm 0.026}$
        & ${\best{0.773}}_{\scriptstyle \pm 0.014}$
        & ${\best{0.598}}_{\scriptstyle \pm 0.010}$
        & ${\second{0.484}}_{\scriptstyle \pm 0.004}$
        & ${\best{60.23}}_{\scriptstyle \pm 0.67}$ \\
        
        \midrule
        \multicolumn{6}{c}{\textbf{Ablation of Exposure-aware Compensation Strategy}} \\
        \midrule
        
        w/o Guidance
        & ${2.399}_{\scriptstyle \pm 0.120}$ 
        & ${0.615}_{\scriptstyle \pm 0.033}$ 
        & ${0.493}_{\scriptstyle \pm 0.008}$ 
        & ${0.455}_{\scriptstyle \pm 0.012}$ 
        & ${59.65}_{\scriptstyle \pm 0.12}$ \\
        
        w/o RFL
        & ${\second{2.986}}_{\scriptstyle \pm 0.032}$ 
        & ${0.632}_{\scriptstyle \pm 0.004}$ 
        & ${\second{0.550}}_{\scriptstyle \pm 0.003}$ 
        & ${\best{0.490}}_{\scriptstyle \pm 0.003}$ 
        & ${58.03}_{\scriptstyle \pm 0.16}$ \\
        
        w/o Clean Ref.
        & ${2.368}_{\scriptstyle \pm 0.031}$ 
        & ${\second{0.656}}_{\scriptstyle \pm 0.018}$ 
        & ${0.514}_{\scriptstyle \pm 0.015}$ 
        & ${0.455}_{\scriptstyle \pm 0.004}$ 
        & ${\second{60.47}}_{\scriptstyle \pm 0.40}$ \\
        
        Rule Prior
        & ${2.276}_{\scriptstyle \pm 0.067}$ 
        & ${0.595}_{\scriptstyle \pm 0.021}$ 
        & ${0.513}_{\scriptstyle \pm 0.008}$ 
        & ${0.469}_{\scriptstyle \pm 0.006}$ 
        & ${59.70}_{\scriptstyle \pm 0.31}$ \\
        
        EPOFusion
        & ${\best{3.171}}_{\scriptstyle \pm 0.008}$
        & ${\best{0.782}}_{\scriptstyle \pm 0.005}$
        & ${\best{0.601}}_{\scriptstyle \pm 0.007}$
        & ${\second{0.482}}_{\scriptstyle \pm 0.003}$
        & ${\best{60.78}}_{\scriptstyle \pm 0.27}$ \\
        
        \bottomrule
    \end{tabular}
    \vspace{-0.3cm}
\end{table}

\subsection{Ablation Study}
\textcolor{red}{To analyze EPOFusion, we conduct ablations on the progressive fusion framework and exposure-aware compensation strategy, reporting fusion metrics and mIoU. All experiments are repeated on IVOE with seeds 0, 1, and 2, as summarized in~\autoref{AblationTable}.}
\subsubsection{Progressive Fusion Framework Ablation}
\begin{figure*}[t]
    \centering
    \includegraphics[width=1.0\textwidth]{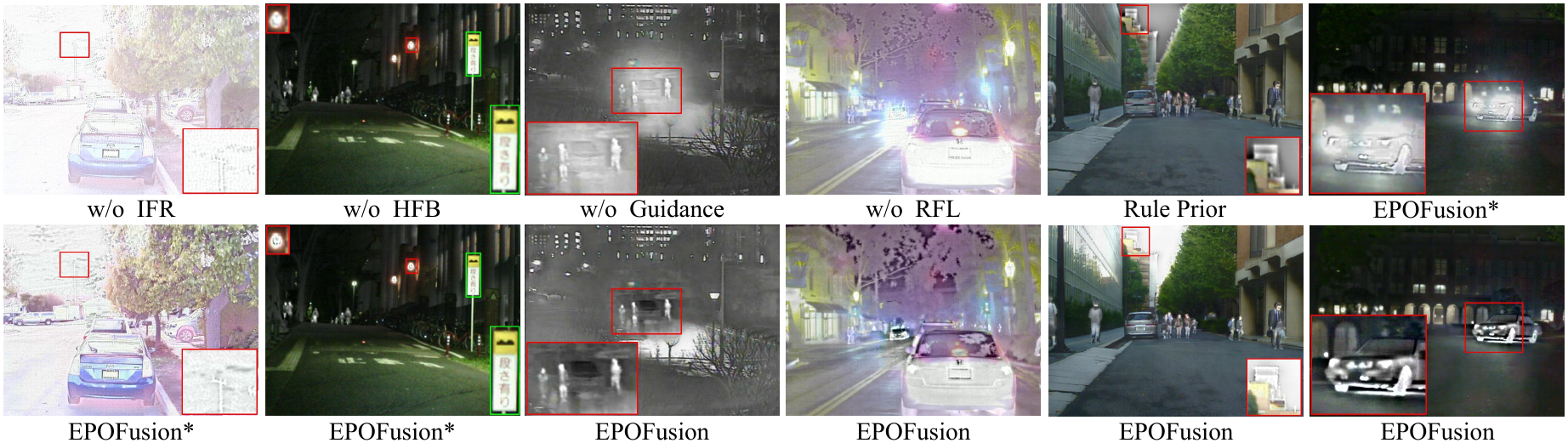}
    \caption{Visual results of the ablation study on EPOFusion* and EPOFusion.}
    \label{AblationPic}
    \vspace{-0.3cm}
\end{figure*}

\textcolor{red}{We first use EPOFusion$^{*}$ as the reference configuration to analyze the key designs of the progressive fusion framework, including iterative feature refinement (IFR), the high-fidelity branch (HFB), and time-dependent encoding (TDE). As shown in~\autoref{AblationTable}, removing IFR reduces VIF, $Q^{AB/F}$, and mIoU from 0.773, 0.598, and 60.23\% to 0.641, 0.548, and 58.07\%, respectively, indicating that iterative refinement progressively integrates complementary multimodal information. As illustrated in~\autoref{AblationPic}, without IFR, structures and local details in overexposed regions become less distinct, whereas the complete progressive fusion framework preserves clearer object contours and texture information. Removing HFB further decreases VIF and $Q^{AB/F}$ to 0.533 and 0.468, respectively. The corresponding visual results also show that HFB helps preserve local textures and details from the source images during progressive reconstruction. Without TDE, VIF, $Q^{AB/F}$, and mIoU decrease to 0.659, 0.554, and 56.76\%, respectively, showing that temporal conditioning helps the network distinguish different refinement stages and adapt feature updates accordingly, thereby promoting the progressive integration of complementary multimodal information. Overall, IFR, HFB, and TDE improve the fused representation from the perspectives of progressive optimization, detail preservation, and stage-aware refinement, respectively.}
\subsubsection{Exposure-aware Compensation Strategy Ablation}
\textcolor{red}{We further use EPOFusion as the reference configuration to analyze the exposure-aware compensation strategy, including the guidance mechanism, region-aware fusion loss (RFL), and clean state reference. As shown in~\autoref{AblationTable}, removing the guidance mechanism reduces VIF, $Q^{AB/F}$, and mIoU from 0.782, 0.601, and 60.78\% to 0.615, 0.493, and 59.65\%, respectively. The guidance mechanism uses the compensation map to locate regions requiring infrared compensation and modulates the progressive state update, allowing feature refinement to focus on locations with reliable infrared structures. As shown in~\autoref{AblationPic}, EPOFusion preserves clearer target structures in overexposed regions. Removing RFL decreases VIF, $Q^{AB/F}$, and mIoU to 0.632, 0.550, and 58.03\%, respectively. The qualitative results show that RFL alleviates overexposure responses while enhancing infrared texture details. The clean feature-state reference provides a stable feature-space target for progressive refinement, contributing to improvements in MI, VIF, and $Q^{AB/F}$. Moreover, unlike the Rule Prior, which uniformly guides the entire overexposed region, our guidance further identifies regions with reliable infrared structures, avoiding redundant infrared background information in the sky while preserving useful roof structures. Compared with EPOFusion$^{*}$, EPOFusion further suppresses overexposure and preserves vehicle structures and background details, demonstrating that the exposure-aware components further enhance infrared information with compensation value on top of the progressive fusion framework.}
\subsubsection{Iteration Step Analysis}
\begin{figure}[t]
    \centering
    \begin{subfigure}[t]{0.425\textwidth}
        \centering
        \includegraphics[width=\linewidth]{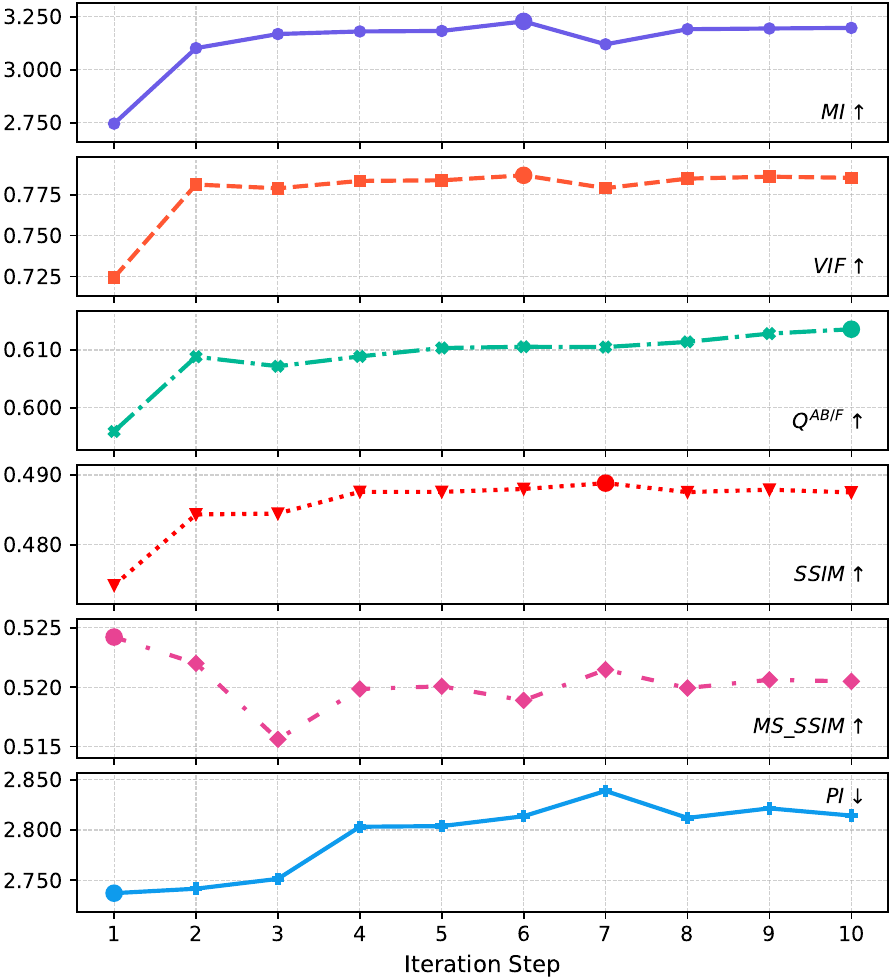}
        \caption{Quantitative comparison.}
        \label{timestep_quantitative}
    \end{subfigure}
    \hfill
    \begin{subfigure}[t]{0.470\textwidth}
        \centering
        \includegraphics[width=\linewidth]{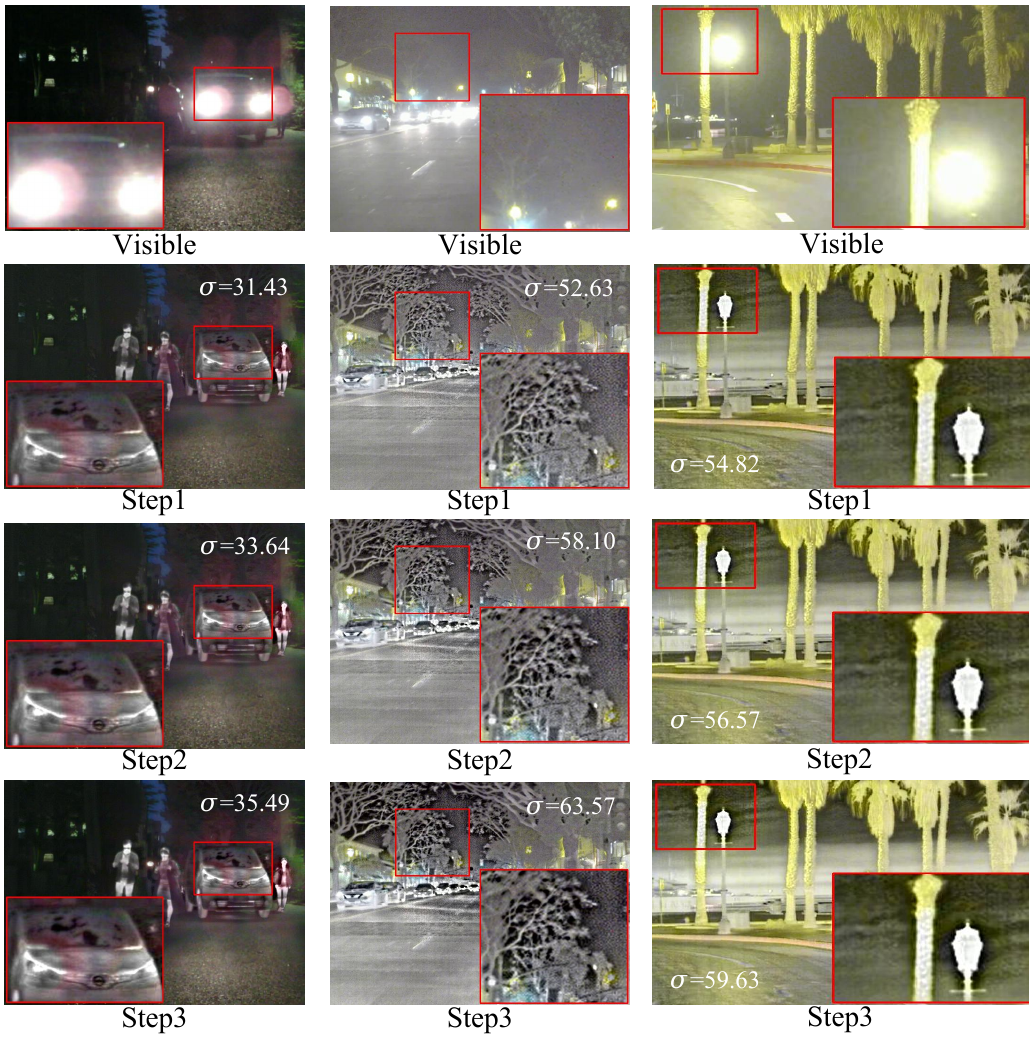}
        \caption{Representative results.}
        \label{timestep_qualitative}
    \end{subfigure}
    \caption{Quantitative and qualitative analysis of EPOFusion with different iteration steps on the IVOE dataset. $\sigma$ denotes the standard deviation of grayscale intensities within the highlighted region.}
    \label{fig:timestep}
    \vspace{-0.3cm}
\end{figure}

\textcolor{red}{We further analyze the effect of the number of iteration steps on fusion performance. As shown in~\autoref{timestep_quantitative}, MI, VIF, $Q^{AB/F}$, and SSIM increase rapidly within the first three iteration steps, after which the gains gradually diminish and tend to stabilize, indicating that most complementary information is effectively integrated during the early refinement stages. The qualitative results further confirm this trend. As shown in~\autoref{timestep_qualitative}, from Step 1 to Step 3, infrared contours in overexposed regions become progressively clearer and texture information becomes richer. In particular, in the second example, vegetation textures gradually emerge as the number of iterations increases; in the third example, the infrared structure around the streetlight is progressively enhanced during the refinement process.}

\section{Conclusion}\label{sec5}
In this paper, we propose EPOFusion, an exposure-aware progressive fusion framework for infrared and visible image fusion under overexposed conditions. To mitigate the insufficient utilization or redundant introduction of infrared information arising from inaccurate perception of infrared details in overexposed regions, EPOFusion introduces an explicit exposure-guided fusion framework to determine what infrared information should be preserved. With infrared-compensation supervision from IVOE, a spatial guidance module identifies regions requiring infrared \textcolor{red}{compensation, while a region-aware fusion loss strengthens informative infrared structures.} In this way, the model can better preserve reliable infrared structures in regions where the visible modality becomes unreliable, while suppressing redundant infrared responses in normally exposed areas.
Moreover, overexposure degrades visible textures and structures, limiting effective infrared preservation in single-step fusion. Therefore, we further design a progressive feature refinement framework equipped with a multi-scale context fusion module. By progressively refining feature representations in degraded regions and integrating local details with global contextual information across different scales, the framework progressively integrates reliable infrared structures in overexposed regions while preserving texture fidelity and visual consistency in normally exposed areas.
\textcolor{red}{We further construct the IVOE dataset, comprising 447 real-world overexposed test pairs with detection and segmentation annotations, to support evaluation under authentic overexposure.} Future work will extend exposure-aware fusion to broader sensing conditions and multimodal perception scenarios.

\section{Author Contributions Statement}
Zhiwei Wang developed the methodology, conducted the experiments, and wrote the original draft of the manuscript. Defeng He contributed to the methodology and reviewed and edited the manuscript. Li Zhao reviewed and edited the manuscript. Xiaoqin Zhang contributed to the methodology. Yuxing Li reviewed and edited the manuscript. Edmund Y. Lam reviewed and edited the manuscript.

\section*{Acknowledgments}
This work was supported partly by the National Natural Science Foundation of China under Grant Nos. U24A20270 and U24A20242, the National Key R\&D Program of China under Grant No. 2024YFC3306901, and the Zhejiang Province Leading Geese Plan under Grant No. 2025C02013.
\bibliographystyle{unsrt}
\bibliography{refs}

\end{document}